\pdfoutput=1 

\RequirePackage{fix-cm}
\documentclass{svjour3}                     
\smartqed  
\usepackage{graphicx} 
\usepackage{float}
\usepackage{amsmath,amssymb,graphicx} 
\usepackage{vwcol} 
\usepackage{lipsum} 
\usepackage{subfig}
\usepackage{algorithm}
\usepackage{algorithmic}
\floatname{algorithm}{Procedure}

\begin{document}

\title{Fast non parametric entropy estimation for spatial-temporal saliency method
}
\subtitle{On general and specific driving conditions}


\author{
Anh Cat Le Ngo \and Guoping Qiu \and Geoff Underwood \and Li-Minn Ang \and Kah Phooi Seng 
}



\date{Received: date / Accepted: date}

\maketitle

\begin{abstract}
This paper formulates bottom-up visual saliency as center surround conditional
entropy and presents a fast and efficient technique for the computation of such
a saliency map. It is shown that the new saliency formulation is consistent
with self-information based saliency, decision-theoretic saliency and Bayesian
definition of surprises but also faces the same significant computational
challenge of estimating probability density in very high dimensional spaces
with limited samples. We have developed a fast and efficient nonparametric
method to make the practical implementation of these types of saliency maps
possible. By aligning pixels from the center and surround regions and treating
their location coordinates as random variables, we use a k-d partitioning
method to efficiently estimating the center surround conditional entropy. We
present experimental results on two publicly available eye tracking still image
databases and show that the new technique is competitive with state of the art
bottom-up saliency computational methods. We have also extended the technique
to compute spatiotemporal visual saliency of video and evaluate the bottom-up
spatiotemporal saliency against eye tracking data on a video taken onboard a
moving vehicle with the driver's eye being tracked by a head mounted
eye-tracker.
\end{abstract}

\def\etal{\emph{et al.}}

\section{Introduction}
\label{sec:intro}

In recent years, there has been increasing interest in the application of
the visual saliency mechanism to computer vision problems. A predominant theory
of the computational visual saliency models is the "center-surround" mechanism
that is ubiquitously found in the early stages of biological vision \cite{HUBEL1965}.
In the literature, a number of computational
models that are in one way or another based on such center-surround theory have
been proposed. Arguably one of the most popular models is that of Itti and Koch
\cite{itti1998model} which computes saliency of a location based on the
difference of the low-level features between the center and its surrounds. Many
variants of this model that directly compute the low-level center surround
differences have subsequently appeared in the literature including those
perform the computation in the frequency domain such as \cite{Hou2007} and
\cite{Guo2008}. An approach that is based on information theory and captures
the center surround differences through self-information of a location in the
context of a given image or a group of other images has been proposed by Bruce
and Tsotos \cite{Bruce2006b}.  Another model that is based on decision theory
has been proposed by Gao \etal \cite{Gao2007} which computes the saliency of a
location through the mutual information between the low- and intermediate-level
features of the location and its surrounds. From a computational perspective,
\cite{Gao2007} and \cite{Bruce2006b} share some similarities in the sense that
whilst \cite{Bruce2006b} computes self-information \cite{Gao2007} computes
mutual information. Another approach that takes a radically different
strategy is that of  Judd \etal \cite{Judd2009} who proposed a learning from
example ideas to compute visual saliency.

A major difficulty with models that are based on the computation of information
quantities such as self-information \cite{Bruce2006b} and mutual information
such as \cite{Gao2007} is that they are computationally challenging - involving
estimating probability density functions in very high dimensional spaces with
limited samples.
The authors of \cite{Bruce2006b} used independent component analysis (ICA) to
project high
dimensional image patches onto independent subspace such that the
self-information can be estimated through 1-d histograms of the independent
components. Computing the data dependent ICA bases is in itself computational
expensive, although some authors have shown that ICA can be replaced by data
independent transforms such as discrete cosine transform \cite {Qiu2007},
computing the transform itself is also computationally demanding.
The authors of \cite{Gao2007} got around the computational difficulty of
estimating the mutual
information by using a parametric Generalized Gaussian Distribution (GGD) to
approximate the probability densities of band-pass center and surround
features. To do this, they
had to resort to statistics of various features of natural images to estimate
the parameters of the GGD's. However, there exists some uncertainties about the
model, and even the authors acknowledged ambivalence with respect to the
importance of the shape parameter \cite{Gao2007}.

In this paper, we follow the center surround principle of visual saliency and
directly formulate the bottom-up saliency of a location as the conditional
entropy of the center given its surround. As the conditional entropy
measures the remaining entropy (or uncertainty, informativeness, or surprise)
of the center after observing the surrounds, this formulation is related to
previous
saliency models of self-information \cite{Bruce2006b}, mutual information
\cite{Gao2007} and Bayesian surprise \cite{Itti2006}. The center
surround conditional entropy formulation gives a direct and intuitive
interpretation of the center surround mechanism and also facilitates the easy
extension to spatio-temporal saliency. As with related models, a significant
challenge is the estimation of probability density functions in high
dimensional spaces with limited samples. A major contribution of this paper is
the development of a
fast and practical solution based on non-parametric multidimensional k-d tree
entropy estimation \cite{Stowell2009a} to make such kinds of approaches
computationally
tractable and make them applicable in real-time applications such as saliency
detection in videos. We present experimental
results on several publicly available eye-tracking still image databases
\cite{Bruce2006b,Judd2009} and a video taken on-board a car with the driver's
eye movement being tracked with a head-mounted eye tracker to demonstrate the
effectiveness of the proposed method and compare it with existing techniques.

\section{Saliency based on Center-Surround Conditional Entropy and Kullback-Leibler Divergence}
\label{sec:method}
	
Let $ I_c(\emph{x},\emph{y})$ be an image patch at location
$(\emph{x},\emph{y})$ and $I_{sr}(\emph{x},\emph{y})$ its surrounding regions.
The conditional entropy of the center given its surround can be defined as
\begin{equation}\label{conditional_entropy}
    \emph{\textbf{H}}(I_c(\emph{x},\emph{y}) | I_{sr}(\emph{x},\emph{y}))=
\emph{\textbf{H}}(I_c(\emph{x},\emph{y}),
I_{sr}(\emph{x},\emph{y}))-\emph{\textbf{H}}(I_{sr}(\emph{x},\emph{y}))
\end{equation}
and can be further defined in terms of joint and marginal probabilities
\begin{equation}\label{conditional_entropy_2}
    \emph{\textbf{H}}(I_c(\emph{x},\emph{y}) | I_{sr}(\emph{x},\emph{y}))=
\sum_{\substack{I_c(\emph{x},\emph{y})\in\emph{\textbf{I}}\\
    I_{sr}(\emph{x},\emph{y})\in\emph{\textbf{I}}}}p(I_c(\emph{x},\emph{y}),I_{sr}(\emph{x},\emph{y}))log\frac{p(I_{sr}(\emph{x},\emph{y}))}{p(I_c(\emph{x},\emph{y}),I_{sr}(\emph{x},\emph{y}))}
\end{equation}
The conditional entropy $\emph{\textbf{H}}(I_c(\emph{x},\emph{y}) |
I_{sr}(\emph{x},\emph{y}))$ can be understood in a number of ways. From a
coding or information theory's perspective, it will take
$\emph{\textbf{H}}(I_c(\emph{x},\emph{y}), I_{sr}(\emph{x},\emph{y}))$ bits to
code the center and its surrounds together, but if we knew the surround
$I_{sr}(\emph{x},\emph{y})$ already, we will have gained
$\emph{\textbf{H}}(I_{sr}(\emph{x},\emph{y}))$ bits of information, and the
conditional entropy measures the remaining bits needed to code the center. From
an uncertainty or informativeness point of view, the conditional entropy
measures the remaining uncertainty of the center once its surrounds are known,
or the amount of information of the center given the knowledge of its
surrounds. We can use the conditional entropy or Kullback-Leibler divergence as a measure of saliency, i.e.
\begin{equation}\label{EntropySaliency}
    \emph{\textbf{S}}(\emph{x},\emph{y})=
\emph{\textbf{H}}(I_c(\emph{x},\emph{y}) | I_{sr}(\emph{x},\emph{y})) \text{ or } \emph{\textbf{S}}(\emph{x},\emph{y})=
\emph{\textbf{D}}(I_c(\emph{x},\emph{y}) | I_{sr}(\emph{x},\emph{y})) 
\end{equation}
The definition of saliency in equation (\ref{EntropySaliency}) and
(\ref{conditional_entropy_2}) is consistent with a number of definitions in the
literature including self-information \cite{Bruce2006b}, surprise
\cite{Itti2006} and decision theoretic saliency \cite{Gao2007}. The
self-information saliency of \cite{Bruce2006b} measures the self-information of
$I_c(\emph{x},\emph{y})$ in the context of its surrounds, $-log\{
p(I_c(\emph{x},\emph{y}))\}$. If $I_c(\emph{x},\emph{y})$ is a common patch
within the image, then $p(I_c(\emph{x},\emph{y}))$ is large, $-log\{
p(I_c(\emph{x},\emph{y}))\}$ will be small, hence the saliency will be small.
$\emph{\textbf{S}}(\emph{x},\emph{y})$ in (\ref{EntropySaliency}) has the same
property, that is, if the center and its surrounds are very similar, then
$\emph{\textbf{S}}(\emph{x},\emph{y})$ will be small and conversely, if they
are very different, it will be large. The surprise measure of
\cite{Itti2006} can be re-written as
\begin{equation}\label{conditional_entropy_surpriseness}
  \emph{\textbf{S}}=
\sum_{\substack{I_c(\emph{x},\emph{y})\in\emph{\textbf{I}} \\
    I_{sr}(\emph{x},\emph{y})\in\emph{\textbf{I}}}}p(I_{sr}(\emph{x},\emph{y}))log\frac{p(I_{sr}(\emph{x},\emph{y}))}{p(I_{sr}(\emph{x},\emph{y})|I_c(\emph{x},\emph{y}))}
\end{equation}
Here, the surrounds $I_{sr}(\emph{x},\emph{y}))$ can be interpreted as the
model or background information and the center $I_c(\emph{x},\emph{y})$ as the
new observation data.  Comparing to our formulated definition of center-surround entropy in terms of Kullback-Leibler divergence,
\begin{equation}\label{kullbackleibler_divergence}
    \emph{\textbf{D}}(I_c(\emph{x},\emph{y}) | I_{sr}(\emph{x},\emph{y}))=
\sum_
	{\substack{I_c(\emph{x},\emph{y})\in\emph{\textbf{I}}\\
           I_{sr}(\emph{x},\emph{y})\in\emph{\textbf{I}}}}
p(I_c(\emph{x},\emph{y}))log\frac{p(I_{sr}(\emph{x},\emph{y}))}{p(I_c(\emph{x},\emph{y}))}
\end{equation}
There is a great similarity between two types of saliency measurements defined by equations \ref{kullbackleibler_divergence} and \ref{conditional_entropy_surpriseness}, so are their responses to input data. Again, the surprise measure will be small when the center and surround are similar and large when they are different. The decision theoretic discriminant saliency of \cite{Gao2007} boils down to the computation of the mutual information between the center and its surround, and the mutual information and the conditional entropy has the following relation.
\begin{equation}\label{mutualinfo&entropy}
    \emph{\textbf{MI}}(I_c(\emph{x},\emph{y}), I_{sr}(\emph{x},\emph{y}))=
\emph{\textbf{H}}(I_c(\emph{x},\emph{y}) ) -
\emph{\textbf{H}}(I_c(\emph{x},\emph{y}) | I_{sr}(\emph{x},\emph{y}))
\end{equation}
$ \emph{\textbf{MI}}(I_c(\emph{x},\emph{y}), I_{sr}(\emph{x},\emph{y}))$ is the
amount of uncertainty of the center $I_c(\emph{x},\emph{y})$ that is removed
when its surrounds $I_{sr}(\emph{x},\emph{y})$ are known. One way to understand
MI is that it measures how much the surround can tell about the center which is
again consistent with the conditional entropy. A large mutual information means
the surround can tell a lot about the center hence the saliency will be low, so
will be the center surround conditional entropy.

For all these definitions of saliency measure, there is a fundamental challenge
- practical implementation. As all involve the estimation of probability
density functions in very high dimensional spaces with limited data samples,
various simplification
processes have to be used, e.g., \cite{Bruce2006b} employed independent
component analysis and \cite{Gao2007} assumed a parametric GGD model. In the
next section, we introduce a fast non-parametric
method.

\section{Fast Estimation of Conditional Entropy based Saliency and Kullback-Leibler Divergence}
\label{sec:implementation}

Visual data have excessive amount of information, but only some of them is
useful for forming saliency maps. Itti \etal \cite{itti1998model}used low-level
features of intensity, colour and orientation to build several conspicuity maps
which are combined in a linear fashion to generate a saliency map. In
the discriminant saliency map approach, Gao \etal \cite{Gao2007} used wavelet
and Gabor filters to extract band-pass features and modeled these features using
parametric Generalized Gaussian distribution (GGD) to estimate the mutual information between
the center and surround. In this paper, we use medium frequencies features
since studies have shown showed that mid-band frequency data
globally allow the best prediction of attention for many categories of scenes
\cite{Urban2010}.
\begin{figure}
\begin{center}
         \includegraphics[width = \linewidth]{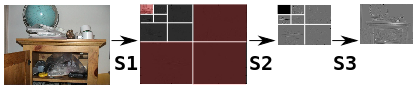}
\end{center}
\caption{Medium Band Filter Flow Chart}
\label{fig:msf1}
\end{figure}
Figure \ref{fig:msf1} shows a step-by-step illustration of mid-band filtering
used in this paper. In the first step, a 9/7 Cohen-Daubechies-Feauveau (CDF) wavelet\cite{Cohen1992} is used to analyse the image into three different frequency bands. After isolating different frequency bands of the visual data, the DC parts and highest frequency-band components, level 3, are removed in step 2. The remaining components are converted back to the image domain by the inverse of the 9/7 CDF wavelet in step 3. DC component is filtered out in order to remove global trend of signals or images while the proposed saliency method mainly depends on center-surround mechanism  or local difference between central and surround patches. Removals of wavelet coefficients level 1 is mainly due to noise reduction purpose. These two steps named as Medium Subband Filter (MSF) is proved to be biologically plausible; Urban \etal \cite{Urban2010} carries out several experiments on correlation between eye fixation data and image features of different frequency bands and concluded that human visual performance well coordinates with image features from wavelet frequency band L2 to L4. More details can be found from Urban paper \cite{Urban2010}. In addition to that good relativity between psychological data and MSF features, another advantage of MSF is cleaning data since it removes DC global trend and noise signals to which the proposed entropy estimation method is very sensible.
\ref{sec:preprocessing}.
\begin{figure}
\begin{center}
         \includegraphics[width=0.5\linewidth]{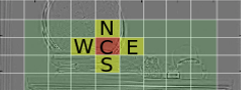}
\end{center}
\caption{Effective Saliency Region}
\label{fig:ent1}
\end{figure}
After medium frequency filtering, the outputs of the step 3 are
divided in 8x8 patches as shown in Figure \ref{fig:ent1}. Although other patch
sizes are possible, 8x8 patches were used in this paper. From those filtered data in patches, saliency values can be estimated as conditional entropy of central data given surrounding data or statistical Kullback-Leibler divergence of data between center and surround patch \ref{EntropySaliency}

First, we consider the saliency as the conditional entropy of each center patch (\textbf{C}) given its four surrounding patches (\textbf{N}, \textbf{S}, \textbf{W}, and \textbf{E}) as shown in Figure \ref{fig:ent1}. The conditional entropy based saliency is computed as
\begin{equation}
    S(\textbf{C}) = H(\textbf{C}|(\textbf{N}, \textbf{S}, \textbf{W},
\textbf{E})) = H(\textbf{C}, \textbf{N}, \textbf{S}, \textbf{W}, \textbf{E}) -
H(\textbf{N}, \textbf{S}, \textbf{W}, \textbf{E})
\label{entropyimplementation}
\end{equation}

Estimating the two joint entropies on the right-hand side of
(\ref{entropyimplementation}) is challenging because of the high dimensionality
of the data. To get round the problem, we take a similar approach as
\cite{Viola1995} and treat the coordinate locations of the pixels as the
random variables and approximate (\ref{entropyimplementation}) as
\begin{equation}\label{entropyimplementation_approximate}
     S(\textbf{C}) = H(c(\emph{x},\emph{y}), n(\emph{x},\emph{y}),
s(\emph{x},\emph{y}), w(\emph{x},\emph{y}), e(\emph{x},\emph{y})) -
H(n(\emph{x},\emph{y}), s(\emph{x},\emph{y}), w(\emph{x},\emph{y}),
e(\emph{x},\emph{y}))
\end{equation}
where $c(\emph{x},\emph{y}), n(\emph{x},\emph{y}), s(\emph{x},\emph{y}),
w(\emph{x},\emph{y}), e(\emph{x},\emph{y})$ are respectively pixels from the
$\textbf{C}, \textbf{N}, \textbf{S}$, $\textbf{W}, \textbf{E}$ patches at the
same reference
location (\emph{x},\emph{y}).

We treat the problem as drawing samples from (\emph{x},\emph{y}) in order to
approximate the conditional entropy. With the formulation of
(\ref{entropyimplementation_approximate}), we can now simplify the problem as
estimating the entropies in the 4-D and 5-D spaces with a total of 8x8 = 64
samples.

We use a technique similar to \cite{Stowell2009a} to achieve fast implementation of (\ref{entropyimplementation_approximate}). The technique is based on a k-d tree style approach to partition the input data space $\Omega\in\Re^D $ into $\emph{A} = \{\emph{A}_j | \emph{j} = 1, 2,
\ldots, \emph{m}\}$ with $\emph{A}_i  \bigcap \emph{A}_j = \varnothing $ if
$\emph{i} \neq \emph{j}$ and $ \bigcup_j \emph{A}_j = \Omega $. Let $
\emph{n}_j $ be the number of samples in the cell $ \emph{A}_j $, $
\emph{V}(\emph{A}_j) $ the volume of cell $ \emph{A}_j $, the total number of
sample $ \emph{N}$, then the multidimensional joint entropy can be estimated as

\begin{equation}\label{kdp_entropy}
    \hat{H}=\sum_{j=1}^{m}\frac{n_j}{N}log\left(\frac{N}{n_j}V(A_j)\right)
\end{equation}

As well as conditional entropy, Kullback-Leibler divergence is another measure which suits into center-surround difference computation. According to the equation \ref{kullbackleibler_divergence}, the saliency of central patch is its divergence in accordance with the surrounding reference. Then, the computational task is figuring out $p(I_{sr}(\emph{x},\emph{y}))$ and $p(I_{c}(\emph{x},\emph{y}))$, where $I_{sr}(\emph{x},\emph{y}) = \{n(x,y),s(x,y),e(x,y),w(x,y)\}$ and $I_{c}(\emph{x},\emph{y}) = \{c(x,y)\}$
 Once again, the technique proposed in \cite{Stowell2009a} comes in handy. The k-d tree partition technique can be utilized to break data points into uniformly distributed cells where probability density function can be simply approximated with number of samples and its correspondent volumes. However, there is one distinguishing point between kdpee usage in conditional entropy and KL divergence estimation which is data ordering. In conditional entropy estimation $H(C|S)$, it does not matter that the joint entropy is computed as $H(C,S)$ or $H(S,C)$. However, that order has importance meaning in KL divergence which is asymmetric information measurement  $ D(C || S) <> D(S || C) $. Hence, before performing k-d partition, data should be organized such that first four dimensions contains data of surrounding patches, and the fifth dimension contains data of central patch. That data format causes surrounding data are partitioned before any central data is done. In other words, surrounding information already  gives reference to the central partitioning tasks. The partition scheme ensures that number of samples in central cells and surrounding cells are equal $n^{sr}_{j} = n^{c}_{j}$, so the ratio $\frac{p(I_{sr}(x,y))}{p(I_{c}(x,y))}$ is equal to $\frac{V(A^{c}_{j})}{V(A^{s}_{j})}$. Therefore, the KL divergence $D(c(x,y) || {n(x,y),s(x,y),e(x,y),w(x,y)})$ in the equation \ref{kullbackleibler_divergence} can be rewritten as.
\begin{equation}\label{kdp_divergence}
    \hat{D}=\sum_{j=1}^{m}\frac{n_j}{N}log\left(\frac{V_{c}(A_j)}{V_{sr}(A_j)}\right)
\end{equation}
The computational complexity of either conditional entropy or KullbackLeibler approach is $\Theta\left(DNlogN\right)$ and the space complexity is $\Theta(DN)$. In our setting, $\emph{N} = 64$ and $\emph{D} = 5$ or $\emph{D} = 4$. The lower limit of the sample size is $\emph{N}\geq2^D$, $2^5 = 32$ and $2^4 = 16$, therefore our setting meets the samples size requirement of the algorithm.

Two measurements and approaches are nearly identical in their meanings, measurements of center-surround information, and computational methods, usage of kd tree partitioning. However, Kullback-Leibler divergence has quite a bit of advantages in computational expenses and preferred visual results. Both boil down to simple truths two kd partitioning jobs need to be done for conditional entropy measurement meanwhile KL divergence only requires once. Therefore, the later approach is essentially twice as fast as the former. In addition, KL divergence reduces the effect of bias in information estimation which are heavily affect saliency maps generated by difference between two entropy estimation. Therefore, it produces much smoother and eye-candy saliency maps. 

Beside better performance, KL divergence possesses interesting theoretical points which help to measure bias in information estimation while conditional entropy does not clearly accounts for it. According to Gibbs' inequality, Kullback-Leibler divergence must be positive; therefore, any negative KL divergence is a result of estimation bias. The bias is mainly due to samples shortage in high-dimensional data estimation; in other words, the bias ratio, number of biased estimation over the total number of estimation, depends on the size of patches, amount of available data for estimation. It paves a way for identifying data size which minimizes possible bias; for example, in 544x720 images, if the odd patch size, $n$, is varied between 7 and 21, we have bias correspondent bias ratio as follows.

\begin{table}[!htbp]
\caption{Bias ration vs. patch size}
\centering
\begin{tabular}{|l|c|c|c|c|c|c|c|c|}
\hline
\textbf{Patch size} & \textit{7x7} & \textit{9x9} & \textit{11x11} & \textit{13x13} & \textit{15x15} & \textit{17x17} & \textit{19x19} & \textit{21x21} \\ \hline
\textbf{Error ratio} & \multicolumn{1}{r|}{0.0352} & \multicolumn{1}{r|}{0.0250} & \multicolumn{1}{r|}{0.0116} & \multicolumn{1}{r|}{0.0046} & \multicolumn{1}{r|}{0.0014} & \multicolumn{1}{r|}{0.0002} & \multicolumn{1}{r|}{0.0000} & \multicolumn{1}{r|}{0.0000} \\ \hline
\end{tabular}
\label{bias_patchsize}
\end{table}

Results in the table \ref{bias_patchsize} points out that a size of patches may needs to reach 19 or 21 to totally eliminate bias in KL divergence estimation. Though 7x7 patch size is responsible for $3.52\%$ more bias estimation points than 21x21 patch size, it does not affect relative saliency value comparison between two points as long as they are estimated on the same number of data. Therefore, it does not causes any significant differences in normalized saliency maps. Choices of size patches does not cause much difference visually and numerically in normalized saliency maps, but it does affect performance in terms of speed. Obviously the more input data, the more processing time needs spending for processing them. In later experiments, 7x7 patches are mainly utilized, since it provides both reasonable saliency maps and computational efficiency.


\section{Extension to Motion Video and Spatiotemporal Saliency}
\label{sec:spatiotemporal}
The techniques described in \ref{sec:method} and \ref{sec:implementation} can
be easily extended to 3D and to compute spatio-temporal saliency maps for videos.
Let $I_c(\emph{x},\emph{y}, \emph{t})$ be the center patch at frame $\emph{t}$,
and $I_{ssr}(\emph{x},\emph{y}, \emph{t})$ its spatial surround defined
similarly as before, 4-neighboring North, South, East and West patches. While $I_{tsr}(\emph{x},\emph{y}, \emph{t})$, the temporal surroundings are central patches of past consecutive temporal frames, $\{I_c(\emph{x},\emph{y}, \emph{t}-1), I_c(\emph{x},\emph{y}, \emph{t}-2), ...,I_c(\emph{x},\emph{y}, \emph{t}-n)\}$. To simplify notation, we will drop the spatial coordinates without causing confusion. The choice of surrounding data leads to reasonable assumption that there is independence between spatial and temporal contexts \cite{Qiu2007}, then we
can define spatio-temporal saliency in term of conditional entropy and Kullback-Leibler divergence as
\begin{equation}\label{conditional_entropy3D}
    \emph{\textbf{S}}_{st}(\emph{t}) = \emph{\textbf{H}}(I_c(\emph{t}) \mid
I_{ssr}(\emph{t})) + \emph{\textbf{H}}(I_c(\emph{t}) \mid I_{tsr}(\emph{t}))
\end{equation}
\begin{equation}\label{kullbackleibler_divergence3D}
    \emph{\textbf{S}}_{st}(\emph{t}) = \emph{\textbf{D}}(I_c(\emph{t}) \mid
I_{ssr}(\emph{t})) + \emph{\textbf{D}}(I_c(\emph{t}) \mid I_{tsr}(\emph{t}))
\end{equation}
The two conditional entropies or KL divergence can be estimated similarly using the technique mentioned in the section \ref{sec:implementation}.
We have applied the spatio-temporal saliency method of
(\ref{conditional_entropy3D}) to process a video taken in the perspective of drivers' eyes
with eyes movement data being recorded with a head-mounted eye tracker. The video and
the eye tracker are synchronized so the driver's fixation points can be simultaneously shown
in the video. The purpose of this experiment is to test the correlation between
computational bottom up visual saliency and the drivers fixation points. A
sample scene and saliency maps corresponding to different saliency methods are
shown in the figure \ref{fig:slm3}. 
\begin{figure}[h]
\begin{center}
         \includegraphics[width = \linewidth]{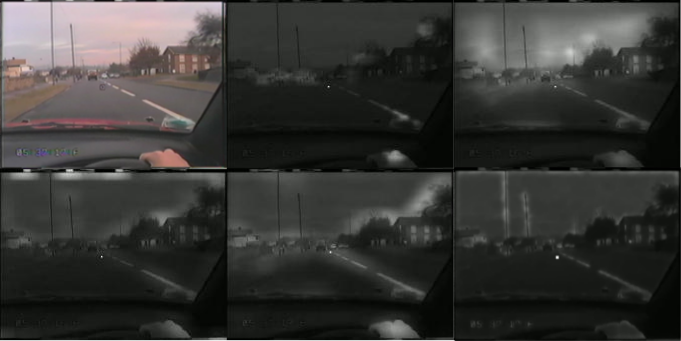}
\end{center}
\caption{Samples and Saliency Maps of ITT, GBVS, PFT, and PQFT, and ENT methods
(left-to-right top-to-bottom order). Please see the supplementary materials for
sample sections of
the video containing the original scene with the eye fixation marker overlayed
on
it and saliency maps of various methods.}
\label{fig:slm3}
\end{figure}
Though the spatial temporal saliency results seems to fit into context of images, the method has not well optimized and analysed yet. Figure \ref{fig:org_ssm_tsm_con_kld}, from left to right, shows original image, spatial saliency maps and temporal saliency maps according to conditional entropy or KL divergence estimation values. In this figure, MSF means medium sub-band feature is in use; SSM and TSM represent spatial saliency maps and temporal saliency map respectively; meanwhile, CON and KLD notifies which kdpee entropy estimation mode in used. First, all saliency maps regardless of approaches ( SSM / TSM or CON / KLD ) identifies edges as high saliency regions. Intuitively, edge regions of images witness sharp change in intensity values between pixels as well as large statistical difference between neighbouring blocks in both spatial and temporal domain. Therefore, surrounding information of edge blocks does not provide much information for identification of central block statistical properties; then, large entropy difference occurs at these edge blocks. Secondly, visual results differentiate performance of KLD and CON approaches; KLD method generates less noise saliency map but less contrasting range; whereas, CON saliency map is much more noisy but the contrast range is larger than the previous method. However, both of them are sensitive to noise, or small change in high intensity area for example sky regions. Besides noise levels in KLD and CON maps, another observable point is high similarity between spatial and temporal saliency maps. The high correlation of pixels between successive frames results in a temporal saliency map (TSM) with much similarity to a spatial saliency map (SSM). Current drawbacks of the proposed method are solely due to its sensitivity to noise and correlation between neighbouring data, and these problem can be solved by de-noising and decorrelation technique later introduced.
\begin{figure}[h]
\begin{center}
         \includegraphics[width = \linewidth]{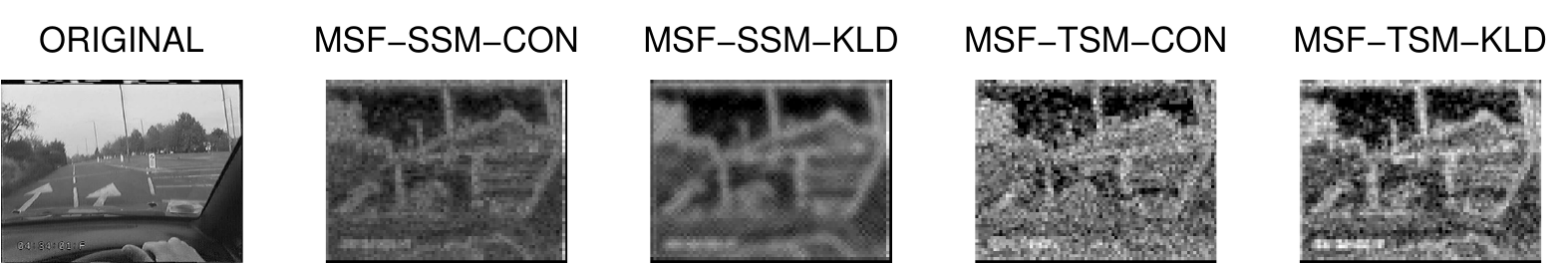}
\end{center}
\caption{Standard Deviation of Each Sub-band in Camera-man Image}
\label{fig:org_ssm_tsm_con_kld}
\end{figure}
\section{Image Patch Data Preprocessing}
\label{sec:preprocessing}

As mentioned in the previous section, the fast entropy estimation method empirically partition data into normal distributed groups by kd-tree methods and compute entropy by simplified entropy formula given the normality of data distribution. Since multi-dimensional data need to be dealt with, adaptive kd-tree is utilized instead of traditional kd-tree partitioning processes. When input data are multidimensional, the advantage of this adaptive process is allowing data to be split along different dimensional at different levels. Then, instead of partitioning n-dimensional data all together, data can be split sequentially one dimension after another. In order to ensure close approximation of adaptive strategy to normal partition, both correlation between data dimensions and number of dimensions itself need to be as small as possible. Therefore in data preprocessing stage, a decorrelation process is crucial for performance of the entropy estimation as well as the the whole entropy-based saliency approach. Beside correlation between multidimensional data, its noise is another factor which sometimes severely affect the entropy estimation performance, so noise removal techniques should be considered before any estimation is taken place. 

\subsection{Data Preprocessing - Dimension Reduction \&\& Decorrelation}
\subsubsection{Principle Component Analysis - Spatial Data Decorrelation}
\label{subsec:pcadec}
In image processing and computer vision research field, Principle Component Analysis (PCA) is well established as de-factor decorrelation tools for data processing because of its reliable theory and good practical performance for short signals of low-dimensional data. Besides decorrelating data, through the analysis, data dimensions can be reduced by eliminating projected low-energy PCA coefficients. It comes into handy in our since the amount of samples desperately limits the number of data dimensions which can be reliably estimated.

\begin{figure}
\begin{center}
         \includegraphics[width = \linewidth]{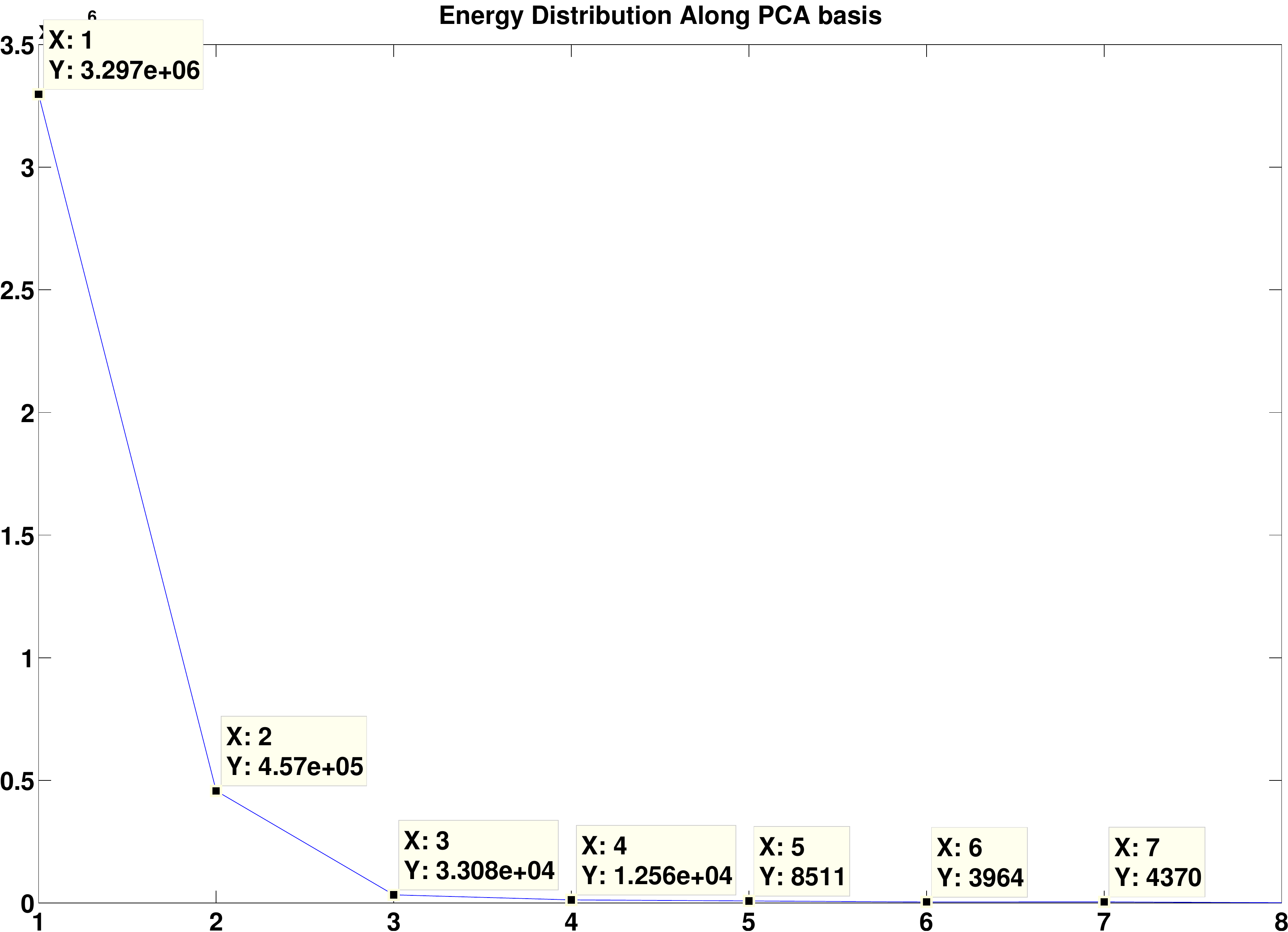}
\end{center}
\caption{Energy Distribution according to PCA basis}
\label{fig:energy_distribution_pca_basis}
\end{figure}

As mentioned in computing approach of spatial saliency map, entropy need to be estimated from high dimensional dataset, the amount of additional entropy for encoding a central patch data given 4-connected surrounding patches, figure \ref{fig:ent1}, need estimating. Conditional approach first carries out estimation of 5-D dataset, $(\textbf{C,N,S,W,E})$, and 4-D dataset, $(\textbf{C,N,S,W,E})$, and then use their difference as saliency values. Since chosen patch size is around 8x8 ( 64 pixels ) or 9x9 ( 81 pixels ), the number of samples are sufficient for 5-D data entropy computation $\mbox(number\_of\_ samples) > 2^5$. However, if 8 connection neighbouring patches are considered, the number of maximum dimensions are no longer 5 but 9. Theoretically, at least $ 2^9 \mbox(pixels/patch)$ or $ 256x256 \text(patches)$ would be used; however it is computationally impractical. Therefore, PCA might be brought in so as to decorrelate and simplify multidimensional data. After the analysis, energy and information of patches are squeezed into first few basis, figure \ref{fig:energy_distribution_pca_basis}. Therefore, ony first few PCA basis or features, 4 or 5 first few basis occupies more than $ 99.8 \%$ of the total energy, can safely describe the pre-analysed data points.

\begin{figure}
\begin{center}
         \includegraphics[width = \linewidth]{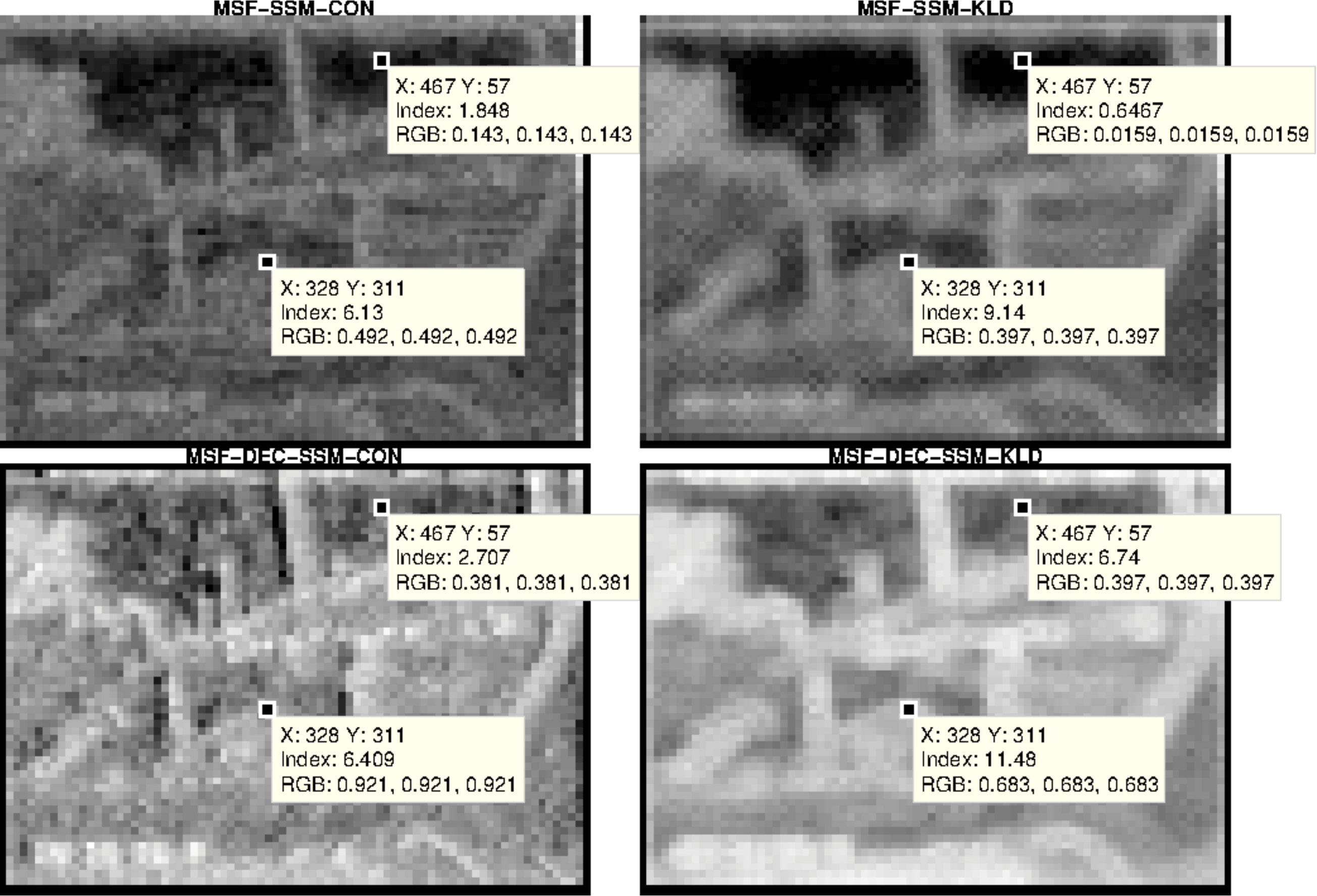}
\end{center}
\caption{Spatial Saliency Maps by CON and KLD with (the bottom row) / without (the top row) DEC (decorrelation)}
\label{fig:dec_vs_non_ssm}
\end{figure}

Despite these manifold data, their underlining structure is much more simpler due to high correlation of natural image patches. If this attribute of neighbouring image patches are taken advantage, the manifold data can be much simplified. For example, the fore-mentioned 9-D input dataset might be simplified into much fewer dimensions meanwhile almost all energy of the original signals is reserved; then, only first 4 or 5 basis are enough to capture almost al information. Hence, 8x8 or 9x9 patch size ( 64 or 81 pixels/patch) provides enough data for 9-D kdpee entropy estimation after PCA process is applied. Data, after decorrelated by PCA, helps the adaptive k-d tree method approximately get close to what normal multidimensional k-d tree method might achieve. A small experiment is carried out to compare between between spatial saliency maps generated by the fore-mentioned method with/without decorrelation, \ref{fig:dec_vs_non_ssm}.

The visual results in the figure \ref{fig:dec_vs_non_ssm} show that decorrelation steps may not be necessary for the proposed saliency method in spatial domain. Saliency values are collected at two points with following coordinates (328,311) ( edge point ) and (467,57) (sky point). At edge points, saliency values are not much different between methods with or without decorrelation steps. However, the values are significant different on supposedly not salient sky point, and they tend to become bigger at those unimportant positions. Visually, decorrelated data generates more noise on the spatial saliency maps. It seems that 4-connecting neighbours provides better context for spatial saliency entropy estimation than 8-connecting neighbours. Therefore, decorrelation step may not be necessary for spatial saliency map generation. This obviously quite a lot of computational effort because averagely it takes 3 times longer to process data with decorrelating step than without it.
\subsection{Discrete Cosine Transform - Temporal Data Decorrelation}
\begin{figure}[!htbp]
\begin{center}
         \includegraphics[width = \linewidth]{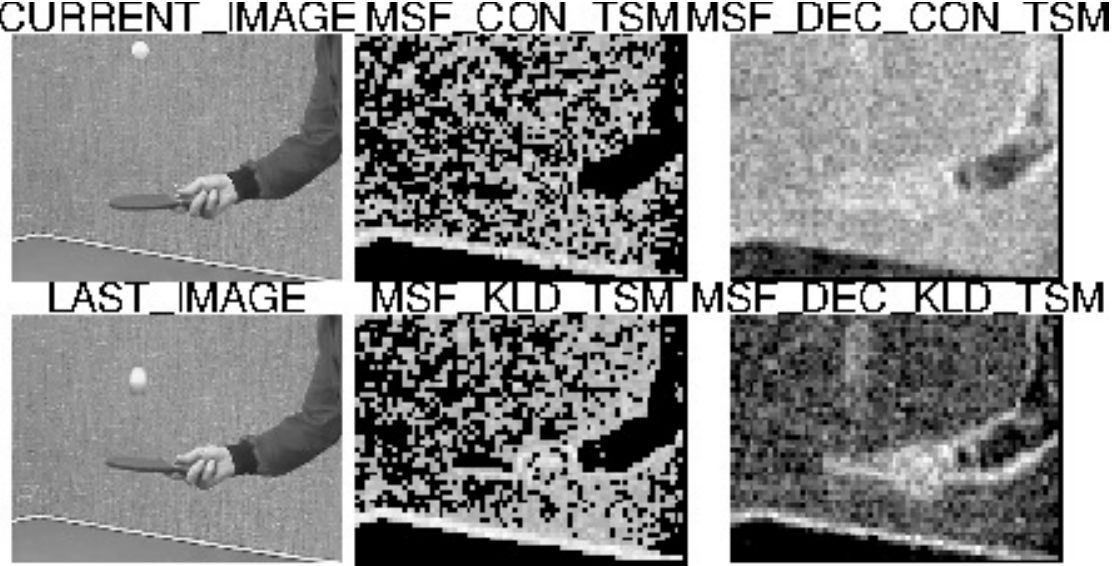}
\end{center}
\caption{Temporal Saliency Map without / with correlation}
\label{fig:dec_vs_non_tsm}
\end{figure}

Though PCA preferred to be used for decorrelating data and reduce data dimension, it is computationally expensive to carry out that analysis repeatedly. Moreover, temporal data need to be dealt with in this section, and Discrete Cosine Transform (DCT) have strong support as fast alternative for decorrelating temporal data instead of PCA in image and video compression. When there are a lot of movements in the scene like two image sample images on relatively noisy background, first column in the figure \ref{fig:dec_vs_non_tsm}. The temporal saliency maps generated by prior temporal approach, the middle column of the figure, clearly show that current method can not cope with those noisy signals and their change over time. There is large temporal correlation between pixels of consecutive frames for both salient foreground features and not salient background features. In addition, if a whole stack of consecutive patches of images without preprocessing is utilized for estimating temporal saliency, it partially covers spatial saliency as well since a whole spatial feature data are in use. Meanwhile, interest of temporal saliency method is emphasizing large movement or big difference in pixel values of video frames. Therefore, in order to decorrelate patch data as well as concentrate on motion features, DCT is employed. Besides its decorrelation mentioned before in video compression literature, its 1D computational processes stress on difference of neighbouring values, in this case difference between pixel values of neighbouring frames. Its coefficients eventually resembles changes of frames over time. After applying DCT analysis on MSF features of several frames, the temporal saliency maps become much clearer and they actually highlight moving objects on the table tennis scenes, right column of the figure \ref{fig:dec_vs_non_tsm}. In addition, the largest coefficients of DCT correlation are removed from temporal saliency estimation in order to limit affects of spatial features on temporal saliency values and noise from background features. Beside removal of the largest component due to its spatial feature,  a few smallest DCT coefficients are as well eliminated so as to reduce data dimensions and complexity while its general structure is still intact and meaningful. For example, 8-D data set, composed from 8 consecutive frames, might be simplified into 3-D by removing the first biggest coefficients and the last four smallest coefficients; then the dataset is much simplified and focused on frame differences. The whole temporal decorrelation process is summarized in pseudo-code \ref{alg:msf3d_decorrelated},\ref{alg:tsm_decorrelated}. Moreover, it is shown that KLD is better than CON when noise is a major part of image signals, but generally temporal saliency maps are still plagued with noisy non-salient parts. Therefore, de-noising, another preprocessing data step, need integrated besides decorrelating process.
\begin{algorithm}
\caption{Medium Sub-band Frequency Filter - Temporal Data Decorrelation}
\label{alg:msf3d_decorrelated}
\begin{algorithmic}
\REQUIRE $input\_frames$
\STATE $msf3d\_filter\_decorrelation(input\_frames)$
\STATE $wavelet\_coefficients = waveletcdf97(input\_frames,number\_wavelet\_level); $
\STATE $wavelet\_coefficients\{L1\} \leftarrow 0;$ \COMMENT{Assign 0 to all L1 wavelet coefficients}
\STATE $wavelet\_coefficients\{DC\} \leftarrow 0;$ \COMMENT{Assign 0 to all DC wavelet component}
\STATE $dct\_coefficients\_reduced = temporal\_decorrelation(wavelet\_coefficients);$
\STATE $msf\_denoised\_feature = \newline inverse\_waveletcdf97(dct\_coefficients\_reduced,number\_wavelet\_level); $
\ENSURE $msf\_decorrelated\_feature$
\end{algorithmic}
\end{algorithm}
\begin{algorithm}
\caption{Temporal Data Decorrelation \& Dimension Reduction}
\label{alg:tsm_decorrelated}
\begin{algorithmic}
\REQUIRE $wavelet\_coefficients$
\STATE \COMMENT{Temporal Data Decorrelation Step}
\FOR {$pixel\_counter=1$ \TO $number\_of\_pixels$}
		\FOR {$frame\_counter = 1$ \TO $number\_of\_frames$}
		\STATE $pixel\_temporal\_data = input\_frames(pixel\_counter,frame\_counter);$
		\ENDFOR
		\STATE $dct\_coefficients(pixel\_counter,:) = dct(pixel\_temporal\_data,:);$
\ENDFOR	
\STATE \COMMENT{Data Dimension Reduction Step}
	\FOR{$pixel\_counter = 1$ \TO $number\_of\_pixels$}
		\FOR{$dct\_basis\_counter = 1$ \TO $number\_of\_dct_basis$}
			\STATE{$basis\_energy(dct\_basis\_counter)  += \{dct\_coefficients(pixel\_counter,dct\_basis_counter)\}^2;$}
		\ENDFOR
	\ENDFOR 	
\STATE \COMMENT{Sort Basis Energy in Descending Order}
	\STATE $basis\_energy\_sorted\_index = sort(basis\_energy,''descend'');$
	\FOR{$pixel\_counter = 1$ \TO $number\_of\_pixels$}
		\FOR{$index\_counter = 2$ \TO $number\_dct\_basis/2$}
			\STATE $dct\_coeffcients\_reduced(pixel\_counter,basis\_energy\_sorted\_index(index\_counter)) \leftarrow dct\_coefficients(pixel\_counter,basis\_energy\_sorted\_index(index\_counter));$
		\ENDFOR
	\ENDFOR
	\ENSURE $dct\_coefficients\_reduced$
\end{algorithmic}
\end{algorithm}
\subsection{Data Preprocessing - Data De-noising}
\subsubsection{Spatial Saliency Map De-noising}
Though in theory MSF features provides suitable data for kdpee estimation due to its freedom from DC signals and noise-rich L1 wavelet features, noise does occurs in other frequency sub-band as well. Therefore, entropy estimation is still sometimes suffered from mixtures  of edges and noise well. It leads to over-estimation of spatial entropy and inaccuracy of saliency values. Beside removal of the highest frequency components from wavelet analysis, further technique must be utilized to suppress unwanted noise in the rest of sub-band without need of removing entire frequency spectrum of images signals. Bivariate shrinkage functions, introduced by Sendur \etal \cite{Sendur2002}, is employed since its de-noising mechanism operates on wavelet domain. In brief, its noise suppression technique bases on correlation between parents and children wavelet coefficients; that correlation implies the wavelet coefficients have high possibility of belonging to salient edges feature if its parent coefficients are extracted from edge features and vice versa. In other words, bivariate shrinkage reserves coefficients which are salient across several scales or sub-band and eliminate noise features which are often only salient in a specific scale. The spatial denoise program is described in pseudo-code \ref{alg:msf2d_denoised}.
\begin{algorithm}
\caption{Medium Subband Frequency Filter - Bivariate Shrinkage De-noising}
\label{alg:msf2d_denoised}
\begin{algorithmic}
	\REQUIRE $input\_image,number\_wavelet\_level$
	\STATE $wavelet\_coefficients = waveletcdf97(input\_image$,$number\_wavelet\_level); $
	\STATE $wavelet\_coefficients\{L1\}  \leftarrow 0;$
	\STATE $wavelet\_coefficients\{DC\} \leftarrow 0;$
	\STATE $denoised\_wavelet\_coefficients = \newline bivariate\_shrinkage$($wavelet\_coefficients,number\_wavelet\_level-1);$
	\STATE $msf\_denoised\_feature = inverse\_waveletcdf97(input\_image,number\_wavelet\_level);$
	\ENSURE $msf\_denoised\_feature$
\end{algorithmic}
\end{algorithm}
\begin{figure}[!htbp]
\begin{center}
         \includegraphics[width = \linewidth]{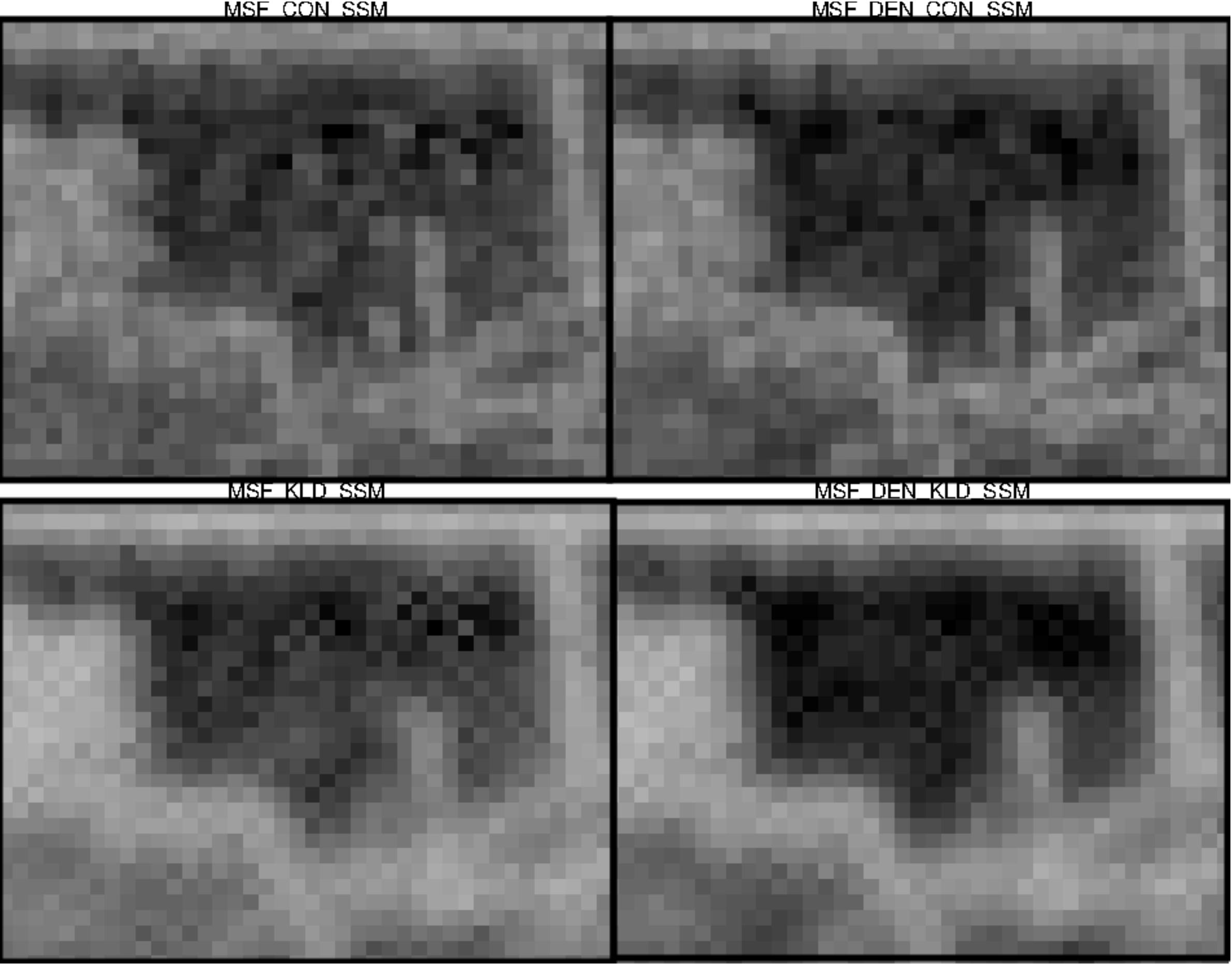}
\end{center}
\caption{Spatial Saliency Map: (left) without denoising, (right) with denoising}
\label{fig:den_vs_non_ssm}
\end{figure}
De-nosing effect on spatial saliency maps can be seen on spatial saliency maps of the figure \ref{fig:den_vs_non_ssm}. It shows saliency maps generated by CON (top row), KLD (bottom row) with de-noising affect ( left column ) and without denoising affect ( right column ).  Its affect can be clearly observed on the bottom row, applying bivariate shrinkage before saliency value KLD estimation technique generates smooth and nearly free noise spatial saliency maps. Especially this de-noising incorporated well into the previous MSF feature extraction because it reuses wavelet CDF 9/7 analysed coefficients. Therefore, it does not requires much addition computational effort.  Due to its good performance and, that bivariate shrinkage might be extended to suppress noisy temporal saliency maps, the figure \ref{fig:dec_vs_non_tsm}
\subsubsection{Temporal Saliency Map De-noising}
Previous sections have already elaborated about how bivariate shrinkage technique removes noise in spatial saliency map. As fore-mentioned, the DCT biggest basis and a few smallest basis are eliminated to avoid spatial features and reduce data complexity. However, their saliency maps are still affected by noise which exist in the rest of DCT basis. Bivariate shrinkage suppress noise on each DCT basis by using correlation between parent and children wavelet coefficients, pseudo-code \ref{alg:msf3d_decorrelated_denoised}.\newline
\begin{figure}[!htbp]
\begin{center}
         \includegraphics[width = \linewidth]{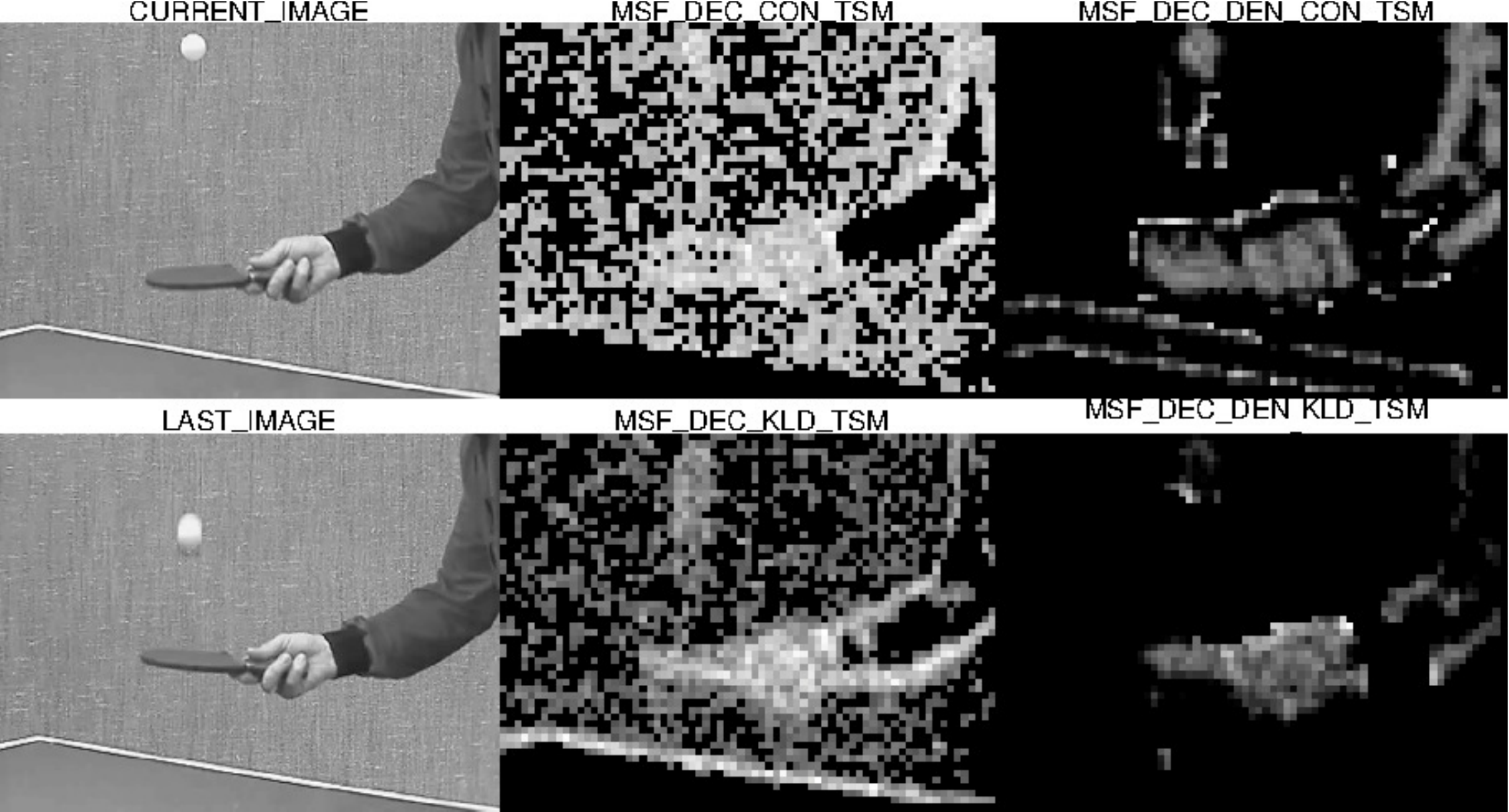}
\end{center}
\caption{Spatial Saliency Map: (left) without denoising, (right) with denoising}
\label{fig:dec_den_vs_dec_tsm}
\end{figure}
Table tennis samples are again used to demonstrate how much noise can be removed from temporal saliency maps after bivariate shrinkage approach. Almost all background noise have been removed from the saliency maps as illustrated in the left-most column of the figure 9 \ref{fig:dec_den_vs_dec_tsm}. Both CON and KLD successfully highlights supposedly saliency objects on the scene which are moving table tennis balls, and arms of players.
\begin{algorithm}
\caption{Medium Sub-band Frequency Filter - Temporal Data Decorrelation \& Denoisation}
\label{alg:msf3d_decorrelated_denoised}
\begin{algorithmic}
\REQUIRE $input\_frames$
\STATE $msf3d\_filter\_decorrelation(input\_frames)$
\STATE $wavelet\_coefficients = waveletcdf97(input\_frames,number\_wavelet\_level); $
\STATE $wavelet\_coefficients\{L1\} \leftarrow 0;$
\STATE $wavelet\_coefficients\{DC\} \leftarrow 0;$
\STATE $dct\_coefficients\_reduced = temporal\_decorrelation(wavelet\_coefficients);$
\FOR{$dct\_basis\_counter = 1$ \TO $number\_of\_dct\_basis\_reduced$}
	\STATE $dct\_coefficients\_reduced\_denoised = \newline bivariate\_shrinkage(dct\_coefficients\_reduced,number\_wavelet\_level-1);$
\ENDFOR
\STATE $msf\_denoised\_feature = \newline inverse\_waveletcdf97(dct\_coefficients\_reduced,number\_wavelet\_level); $
\ENSURE $msf\_decorrelated\_feature$
\end{algorithmic}
\end{algorithm}
\section{Experimental Results}
\label{sec:result}
We evaluate the new conditional entropy based saliency method (from now on
referred to as ENT) on publicly available eye tracking databases of Bruce and Tsotos \cite{Bruce2006b} and Judd \etal \cite{Judd2009}, and compare it with a number of saliency estimation methods in the literature including, Itti and Koch (ITT) \cite{itti1998model}, spectral residual saliency (SRS) \cite{Hou2007}, phase frequency transform (PFT) and phase quaternion frequency saliency (PQFT) \cite{Guo2008}, Information Maximization (AIM)\cite{Bruce2006b}, and discriminant saliency (DIS) \cite{Gao2007}. Among these saliency methods, PQFT and the proposed method incorporate both spatial and temporal information meanwhile, others solely generate saliency maps in frame-by-frame manner. Therefore, experiments on both spatial and spatio-temporal data are in need.
\subsection{Spatial Expeiments}
\label{subsec:spatial_experiment}
In Bruce's image database \cite{Bruce2009a}, eye tracking data were collected
from 20 human subjects over 120 different colour images. Scenes of these photos
include both outdoor and indoor environment, some contain very clear and big
salient objects, but some contain a variety of small objects. This image
database has been used by the data originator and a number of other
researchers for evaluating saliency estimation methods.
\begin{figure}[!htbp]
\begin{center}
         \includegraphics[width = \linewidth]{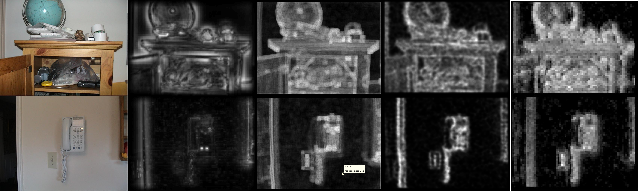}
\end{center}
\caption{Examples of Saliency Maps of ITT, AIM, and the new ENT methods
(left-to-right)}
\label{fig:slm1}
\end{figure}
Figure \ref{fig:slm1} shows the saliency maps of a number of example images
generated by different methods. It is seen that the visual appearance of these
saliency maps are quite similar.

\begin{figure}
\begin{center}
\begin{tabular}{cc}
         \includegraphics[width = 0.49\textwidth]{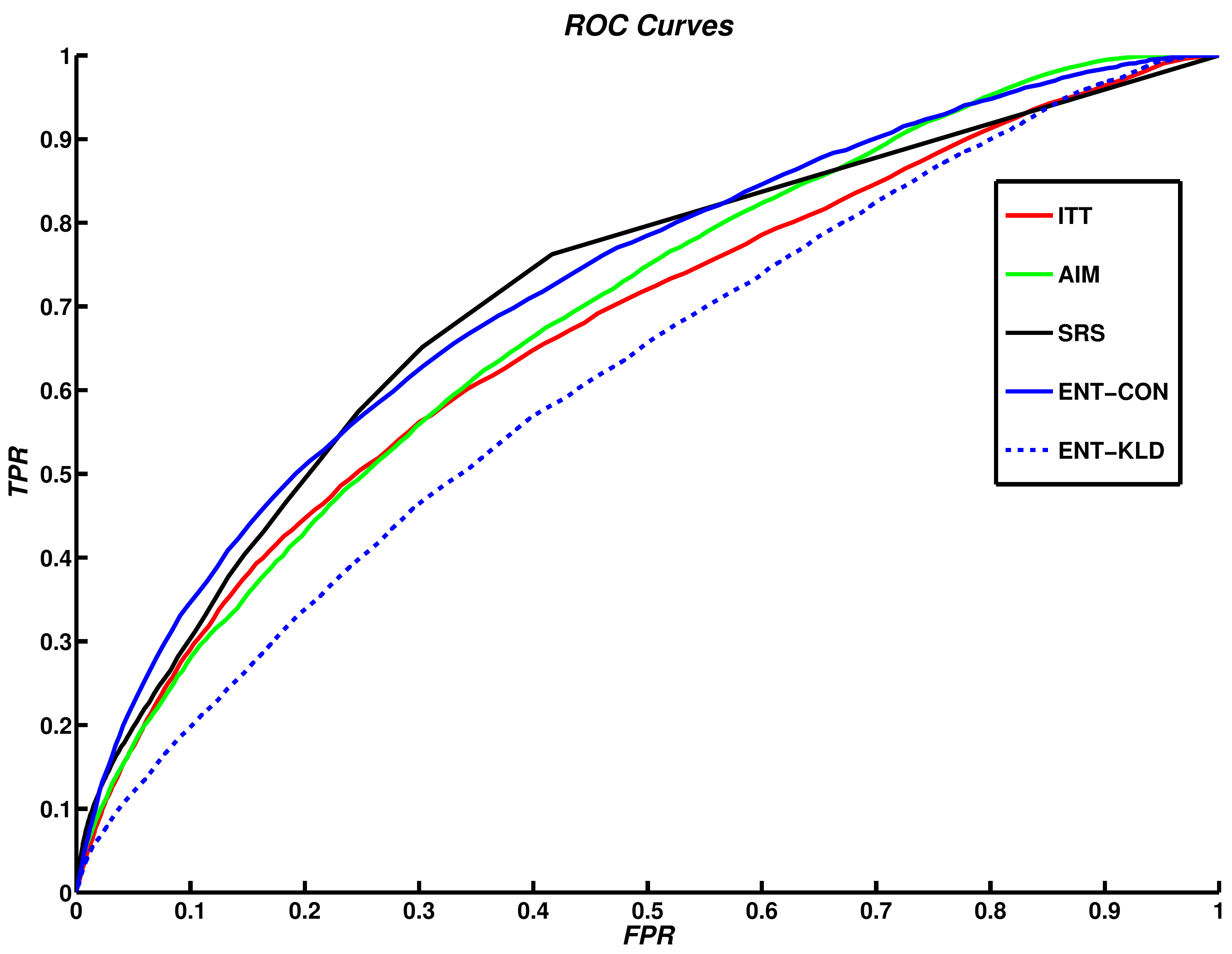}
          &
         \includegraphics[width = 0.49\textwidth]{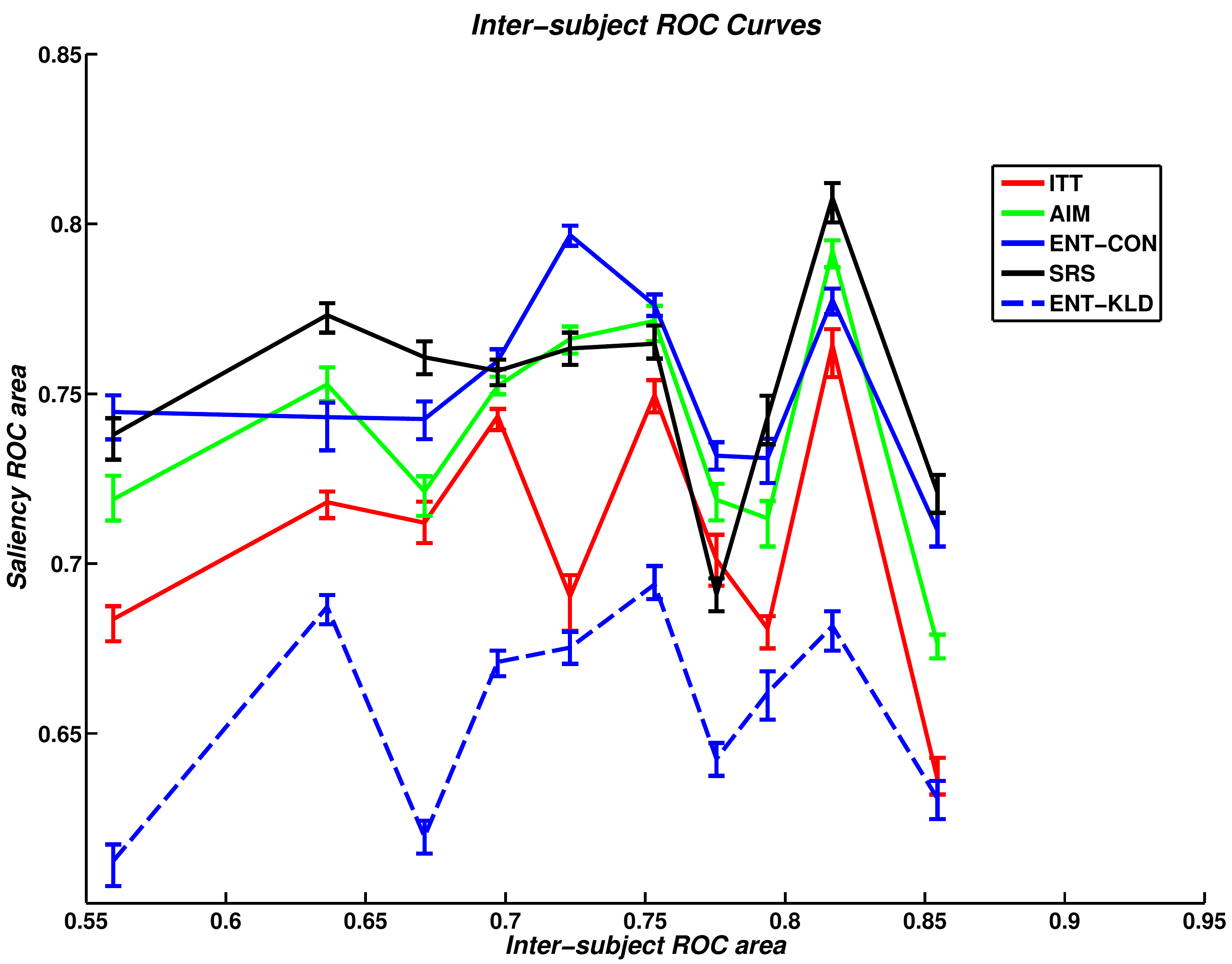}
\end{tabular}
\end{center}
\caption{ROC and Inter-subject ROC curves of ITT, AIM, and ENT-CON/KLD}
\label{fig:roc1}
\end{figure}

To compare the performances of different methods quantitatively, we can use
Tatler's numeric measurement \cite{Tatler2005}. The saliency maps are treated
as binary classifiers to discriminate fixation points versus non-fixation points.
The threshold for classifying fixation points are not fixed but systematically
changed from the minimum to the maximum of the saliency maps to generate ROC
curves. The ROC curves of various methods tested on Bruce's database
\cite{Bruce2009a} are shown in Figure \ref{fig:roc1}. The area under the ROC
curves (AUC) have been used by a number of authors to give quantitative
comparison of saliency computation methods and table \ref{tab:auc1} shows the
AUC values of five different methods.

\begin{table}
\caption{Area Under Curve (AUC) for different methods. }
\begin{center}
\begin{tabular}{|l|l|l|l|l|l|l|}
\hline
Methods & \multicolumn{1}{l|}{ITT \cite{itti1998model}} &
\multicolumn{1}{l|}{AIM \cite{Bruce2006b}} & \multicolumn{1}{l|}{ENT-CON} & \multicolumn{1}{l|}{ENT-KLD} &
\multicolumn{1}{l|}{DIS\cite{Gao2007}} &
\multicolumn{1}{l|}{SRS\cite{Hou2007}}\\ \hline
AUC & 0.70947 & 0.73873 & 0.78167 & 0.73280 & 0.76940 & 0.75434\\ \hline
\end{tabular}
\end{center}
\label{tab:auc1}
\end{table}

The ROC curves show that the new ENT-CON method generally performs better than AIM and ITT methods, and the performances are reconfirmed by the area under curve (AUC) results in table \ref{tab:auc1}. In the table, the AUC result of DIS saliency method was performed on the same database by the original authors and taken directly from \cite{Gao2007}. These AUC results show that ENT
methods also performs better or at least as good as the DIS method. 

ROC curves and AUC values are useful for comparing different computational
saliency approaches, but they do not show relation between these methods with
human performance. The inter-subject ROC curves performance evaluation methods
proposed
by Harel \etal \cite{Harel2007a} helps to show performance of human visual
system versus that of computational saliency method: for each image, a mean
inter-subject ROC area was computed as follows: for each of the subjects who
viewed an image, the fixation points of all other subjects were convolved with
a circular, decaying kernel with decay constant matched to the decaying cone
density in the retina. This was treated as a saliency map derived directly from
human fixations, and with the target points being set to the fixations of the
chosen subject, an ROC area was computed for a single subject. The mean over
all subjects is termed "inter-subject ROC value". For any particular
computational scheme, an ROC area was computed using the resulting saliency map
together with the fixations from all human subjects as target points to detect.
The inter subject ROC values are shown in Figure \ref{fig:roc1}. This plot
clearly demonstrates that our new ENT technique has outperformed current state-of-art
saliency methods and displayed a good matching with human-eye fixation points.

\begin{table}
\caption{Time Consumption of Saliency Methods}
\begin{center}
\begin{tabular}{|c|c|c|c|c|c|}
\hline
Methods & {ITT} & {AIM} & {ENT-CON} & {ENT-KLD} & {SRS} \\ 
\hline
Time (s) & 1.2488 & 66.2673 & 0.93094 & 0.56824 & 0.33654\\ 
\hline
\end{tabular}
\end{center}
\label{tab:tim1}
\end{table}

Information-theoretic saliency methods such as AIM has disadvantages due to its
massive computational requirements which are not suitable for realtime
applications. Therefore, in addition to an accuracy requirement, the proposed
method aims to achieve computational efficiency. Table \ref{tab:tim1} shows the
computational speeds of several techniques, it is seen that ENT is over 70
times faster than AIM method and 1.3 times faster than ITT method, noted that only the spatial map of the proposed method is generate to be fair in speed comparisons with other frame-by-frame based saliency method. Though it is
slower than SRS method, ENT can better match eye-fixation data than SRS. All
experiments are done using MATLAB 2010b in a 2.33 GHz Intel Core 2 Duo computer
running Linux Ubuntu 10.10 OS.
\begin{table}
\caption{Area Under Curve (AUC)}
\begin{center}
\begin{tabular}{|l|l|l|l|l|}
\hline
Methods & \multicolumn{1}{l|}{ITT} & \multicolumn{1}{l|}{AIM} &
\multicolumn{1}{l|}{ENT-CON} & \multicolumn{1}{l|}{MIT\cite{Judd2009}} \\ \hline
AUC & 0.74940 & 0.71165 & 0.78157 & 0.68845 \\ \hline
\end{tabular}
\end{center}
\label{tab:auc2}
\end{table}

\begin{figure}
\begin{center}
         \includegraphics[width = \linewidth]{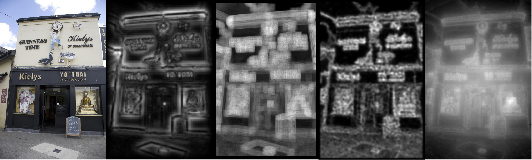}
\end{center}
\caption{Samples and Saliency Maps of ITT,  AIM, ENT, and MIT 
methods (left-to-right order)}
\label{fig:slm2}
\end{figure}

We have also tested our method on the database created by Judd \etal
\cite{Judd2009}. This database has 1003 photos with eye fixation data from 15
viewers, and the database was used for evaluating Judd's learning based
saliency method. We carried out similar qualitative and quantitative evaluation
as \cite{Judd2009} on the 100 testing images used in the original paper. Visual
illustration of a sample photo and its saliency maps of ITT, AIM, ENT and
MIT\cite{Judd2009} is shown in the figure \ref{fig:slm2}. Quantitative results
are shown in the table \ref{tab:auc2} which again show that the new ENT-CON
method compares very well against other methods.

\subsection{Spatiotemporal Experiments}
\label{subsec:spatiotemporal_experiment}
Aforementioned experiments only focus on current frame information or spatial side of available information; however, the proposed ENT method  is naturally extended to estimating saliency from temporal information of available data as well, introduced in the section \ref{sec:spatiotemporal}. For quantitative performance evaluation and comparisons of the spatiotemporal and spatial saliency methods, simulations must be done only spatiotemporal input data; in other words, video materials. Moreover, this paper focuses in specific video context, human visual perception in driving contexts; therefore, the supplementary video data must be recorded in drivers' point of views and psychological related data eye-fixation or ground truth segments done by human beings. Following these criteria, there are two video database put in use, AUTOM and MSDR \cite{Brostow2008}. AUTOM database includes a round 10 minutes videos, recored by a camera on head-mounted eye-trackers which at the same time detect drivers' fixation points on road scenes. Meanwhile, MSDR is as well recorded on road with drivers' view in high resolution but it has ground truth for each frame instead of eye-fixation data. Experiments on AUTOM database emphasizes correlation between artificial saliency maps of different computational method with human psychological data; meanwhile, MSDR with ground truths for specific purposes stresses on specific application how well saliency maps relates to important features in the scenes.
\subsubsection{AUTOM database}
\label{subsubsec:automdb}
The AUTOM video database is created by Accident Research Unit, the University of Nottingham, UK Campus, which includes 28 short segmented videos; each has 500 frames except the last video which has 512 frames. The whole video database has 14012 frames which lasts for 9 minutes and 20 seconds. These videos are recorded with a head-mounted camera of SMI eye-trackers; simultaneously, two small cameras, pointing to pupils of the driver's eyes, locate drivers' eye fixation at 250 Hz. The videos are recorded in real-life driving situations; therefore, several road types and traffic situations have been covered such as urban roads, high ways, speeding up and slowing down situations. It does not extensively covers all possible driving schemes, but it provides sufficient materials about human eye-fixation in most daily driving situations. Importantly, each frame of these videos are recorded with drivers' eye fixation response to specific real-life driving situations. Different from previous ways of collecting multiple eye-fixation data for each static picture, there is only one eye-fixation position data for each frame. The data format is changed, so are evaluation methods since ROC, AUC and ISROC are not suitable for the evaluation of video with eye-marks.

In saliency literature, Normalized Saliency Value (NSV) and Chance Adjusted Saliency (CAS) have been proposed for correlating human performance and generated saliency maps spatiotemporally. 
\begin{table}[!htbp]
\caption{Normalized Saliency Value}
\begin{center}
\begin{tabular}{|l|c|c|c|c|c|c|}
\hline
Observations & \multicolumn{1}{l|}{ITTI} & \multicolumn{1}{l|}{GBVS} & \multicolumn{1}{l|}{PFT} & \multicolumn{1}{l|}{PQFT} & \multicolumn{1}{l|}{ENT-CON} & \multicolumn{1}{l|}{ENT-KLD} \\ \hline
sample\_00 & 0.026 & 0.39 & 0.145 & 0.146 & 0.398 & 0.596 \\ \hline
sample\_01 & 0.018 & 0.339 & 0.131 & 0.136 & 0.318 & 0.444 \\ \hline
sample\_02 & 0.033 & 0.359 & 0.156 & 0.156 & 0.489 & 0.671 \\ \hline
sample\_03 & 0.028 & 0.406 & 0.185 & 0.195 & 0.556 & 0.708 \\ \hline
sample\_04 & 0.01 & 0.306 & 0.099 & 0.1 & 0.242 & 0.348 \\ \hline
sample\_05 & 0.029 & 0.28 & 0.08 & 0.088 & 0.209 & 0.271 \\ \hline
sample\_06 & 0.028 & 0.29 & 0.096 & 0.122 & 0.254 & 0.364 \\ \hline
sample\_07 & 0.008 & 0.371 & 0.125 & 0.164 & 0.267 & 0.39 \\ \hline
sample\_08 & 0.022 & 0.378 & 0.141 & 0.161 & 0.323 & 0.468 \\ \hline
sample\_09 & 0.014 & 0.433 & 0.156 & 0.169 & 0.411 & 0.548 \\ \hline
sample\_10 & 0.002 & 0.306 & 0.077 & 0.096 & 0.145 & 0.24 \\ \hline
sample\_11 & 0.015 & 0.336 & 0.107 & 0.101 & 0.241 & 0.341 \\ \hline
sample\_12 & 0.034 & 0.387 & 0.139 & 0.13 & 0.369 & 0.545 \\ \hline
sample\_13 & 0.015 & 0.38 & 0.142 & 0.134 & 0.338 & 0.435 \\ \hline
sample\_14 & 0.017 & 0.381 & 0.215 & 0.163 & 0.564 & 0.69 \\ \hline
sample\_15 & 0.064 & 0.409 & 0.174 & 0.185 & 0.563 & 0.704 \\ \hline
sample\_16 & 0.024 & 0.257 & 0.082 & 0.101 & 0.216 & 0.353 \\ \hline
sample\_17 & 0.027 & 0.527 & 0.231 & 0.271 & 0.555 & 0.736 \\ \hline
sample\_18 & 0.022 & 0.423 & 0.157 & 0.198 & 0.327 & 0.442 \\ \hline
sample\_19 & 0.065 & 0.286 & 0.06 & 0.103 & 0.159 & 0.24 \\ \hline
sample\_20 & 0.123 & 0.423 & 0.123 & 0.175 & 0.363 & 0.582 \\ \hline
sample\_21 & 0.16 & 0.374 & 0.118 & 0.151 & 0.36 & 0.61 \\ \hline
sample\_22 & 0.059 & 0.341 & 0.099 & 0.157 & 0.282 & 0.456 \\ \hline
sample\_23 & 0.077 & 0.354 & 0.135 & 0.155 & 0.355 & 0.643 \\ \hline
sample\_24 & 0.048 & 0.411 & 0.177 & 0.189 & 0.409 & 0.589 \\ \hline
sample\_25 & 0.069 & 0.389 & 0.194 & 0.192 & 0.475 & 0.695 \\ \hline
sample\_26 & 0.019 & 0.415 & 0.238 & 0.239 & 0.482 & 0.628 \\ \hline
sample\_27 & 0.097 & 0.418 & 0.243 & 0.244 & 0.558 & 0.746 \\ \hline
Samples & 0.041 & 0.37 & 0.144 & 0.158 & 0.365 & 0.517 \\ \hline
\end{tabular}
\end{center}
\label{tab:NSV}
\end{table}
Normalized Saliency Value is in fact collecting saliency values of normalized saliency maps at locations of eye-fixations; the larger NSV is, the more accurate saliency methods can predict human attention points.  However, if the saliency values is directly retrieved at locations of eye-marks, they are prone to a mean spatial error of eye-trackers usually $0.5^{\circ}-1^{\circ}$. So as to eliminate those errors, the maximum saliency value in square areas ,whereof center is the fixation point, is chosen instead. In the table \ref{tab:NSV}, a row $Sample_{\#\#}$ represents normalized saliency values of the $\#\#^{th}$ video in the database with respect to six different saliency approaches. The normalized saliency values in each video are calculated by averaging these of all its frames. Last row shows averagely how much saliency values at eye-fixations are in consideration of the whole 10 minutes videos. Averagely, either conditional entropy or Kullback-Leibler saliency methods outperform the other state-of-the-art approaches. Second to our method in average NVS values is GBVS method, then PQFT, PFT and ITT saliency methods are ranked in descending orders.
\begin{table}[!htbp]
\caption{Chance Adjusted Saliency}
\begin{center}
\begin{tabular}{|l|c|c|c|c|c|c|}
\hline
Observations & \multicolumn{1}{l|}{ITTI} & \multicolumn{1}{l|}{GBVS} & \multicolumn{1}{l|}{PFT} & \multicolumn{1}{l|}{PQFT} & \multicolumn{1}{l|}{ENT-CON} & \multicolumn{1}{l|}{ENT-KLD} \\ \hline
sample\_00 & -0.01620 & 0.15105 & -0.01912 & -0.02068 & 0.09810 & 0.16031 \\ \hline
sample\_01 & -0.02883 & 0.10282 & -0.05274 & -0.05878 & -0.00154 & -0.01592 \\ \hline
sample\_02 & -0.00547 & 0.13573 & -0.03199 & -0.03077 & 0.16424 & 0.20673 \\ \hline
sample\_03 & -0.01498 & 0.19607 & 0.03642 & 0.02633 & 0.27253 & 0.29317 \\ \hline
sample\_04 & -0.03208 & 0.08501 & -0.08824 & -0.08548 & -0.07596 & -0.10103 \\ \hline
sample\_05 & -0.01848 & 0.05986 & -0.11093 & -0.08703 & -0.09415 & -0.15169 \\ \hline
sample\_06 & -0.02129 & 0.07737 & -0.05542 & -0.04822 & -0.03347 & -0.04137 \\ \hline
sample\_07 & -0.02886 & 0.12561 & -0.04424 & -0.04983 & -0.06474 & -0.06572 \\ \hline
sample\_08 & -0.01703 & 0.13734 & -0.04811 & -0.04805 & -0.01329 & 0.01062 \\ \hline
sample\_09 & -0.02898 & 0.18808 & -0.03441 & -0.04354 & 0.04525 & 0.06153 \\ \hline
sample\_10 & -0.04026 & 0.08053 & -0.11124 & -0.12110 & -0.22137 & -0.26291 \\ \hline
sample\_11 & -0.02833 & 0.09366 & -0.09500 & -0.08694 & -0.07029 & -0.10209 \\ \hline
sample\_12 & -0.00883 & 0.16270 & -0.03029 & -0.03776 & 0.07808 & 0.12318 \\ \hline
sample\_13 & -0.02912 & 0.16786 & -0.02083 & -0.02337 & 0.04448 & 0.02625 \\ \hline
sample\_14 & -0.02420 & 0.15050 & 0.01061 & -0.02473 & 0.22266 & 0.22460 \\ \hline
sample\_15 & 0.01803 & 0.17338 & -0.00786 & 0.00165 & 0.24206 & 0.26677 \\ \hline
sample\_16 & -0.01952 & 0.05000 & -0.08519 & -0.07502 & -0.08264 & -0.06527 \\ \hline
sample\_17 & -0.01150 & 0.27638 & 0.03338 & 0.09682 & 0.25085 & 0.28031 \\ \hline
sample\_18 & -0.02229 & 0.16684 & -0.02129 & 0.00881 & 0.03661 & 0.00775 \\ \hline
sample\_19 & 0.01972 & 0.07822 & -0.08941 & -0.07300 & -0.10352 & -0.16421 \\ \hline
sample\_20 & 0.08400 & 0.21136 & -0.01114 & 0.00762 & 0.10156 & 0.16437 \\ \hline
sample\_21 & 0.11576 & 0.20080 & -0.00166 & 0.00948 & 0.12431 & 0.24193 \\ \hline
sample\_22 & 0.01808 & 0.13805 & -0.01966 & -0.01490 & 0.02612 & 0.06633 \\ \hline
sample\_23 & 0.03123 & 0.14159 & -0.01558 & -0.02963 & 0.05672 & 0.20264 \\ \hline
sample\_24 & 0.00010 & 0.22466 & 0.03073 & 0.01936 & 0.14511 & 0.18615 \\ \hline
sample\_25 & 0.02951 & 0.19172 & 0.03718 & 0.03137 & 0.21777 & 0.30256 \\ \hline
sample\_26 & -0.02134 & 0.18658 & 0.07070 & 0.05061 & 0.18509 & 0.19666 \\ \hline
sample\_27 & 0.05362 & 0.21951 & 0.09223 & 0.05795 & 0.28961 & 0.34264 \\ \hline
Samples & -0.00170 & 0.14905 & -0.02440 & -0.02317 & 0.06572 & 0.08551 \\ \hline
\end{tabular}
\end{center}
\label{tab:CAS}
\end{table}
Chance-adjusted saliency performance metric proposed by Parkhurst \etal \cite{Parkhurst2002a} emerged in the literature as the preferred metric for saliency method comparisons. It basically measure difference in correlation of each saliency method between fixation points generated by human visual systems and random number generators. The more significant the difference (CAS) is, the better saliency maps can distinguish between human eye fixations and randomly generated fixations. In the table \ref{tab:CAS}, GBVS saliency method has more success averagely than other saliency methods in distinguishing between human and random eye-fixations. Second to GBVS are our proposed methods ENT-CON and ENT-KLD, then ITTI, PFT and PQFT are ranked in descending order.

Looking at descending orders in performance according to NSV and CAS of saliency methods $R_{NSV} = \{ENT-KLD,ENT-CON,GBVS,PQFT,PFT,ITTI\}$ and $R_{CAS} = \{GBVS,ENT-KLD,ENT-CON,ITTI,PQFT,PFT\}$, we can make conclusions that groups of saliency methods GBVS, ENT-KLD and ENT-CON outperforms the other group PQFT,PFT, and ITTI. However, there is discrepancy inside each group when choosing CAS or NSV as evaluating methods. In a superior group, our proposed method generates saliency maps with more specific local details than GBVS does. That is a reason why ENT-KLD / ENT-CON performs better than GBVS in NSV but worse than GBVS in CAS. In order to make the comparison fairer between local feature saliency methods and their global counterparts, region of interest mask are applied on local saliency maps. The ROI masks are simply generated by setting threshold on saliency values. Similar explanation can be used for discrepancy in CAS, and NSV results of inferior group of saliency methods.

\begin{figure}
  \captionsetup[subfloat]{format=hang,labelfont={rm,bf},textfont={footnotesize},labelformat=simple,labelsep=period,margin=5pt,justification=raggedright}
\centering
  \subfloat[go\_alone]{\label{fig:goAlone}\includegraphics[width = 40mm]{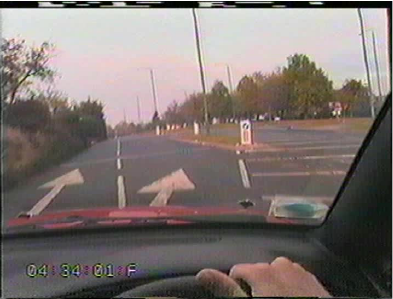}} \qquad
  \subfloat[follow\_car]{\label{fig:followCar}\includegraphics[width = 40mm]{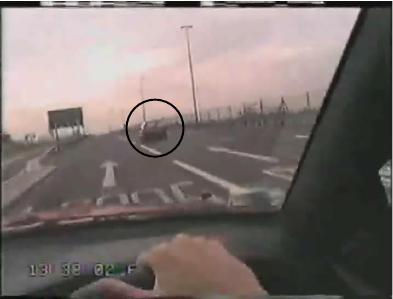}} \\
  \subfloat[slow\_down]{\label{fig:slowDown}\includegraphics[width = 40mm]{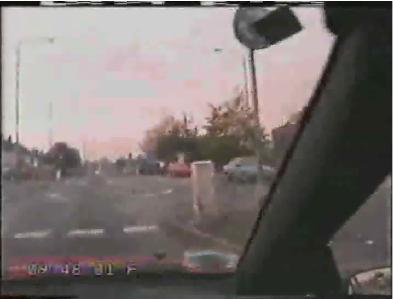}}\qquad  
  \subfloat[go\_toward]{\label{fig:goToward}\includegraphics[width = 40mm]{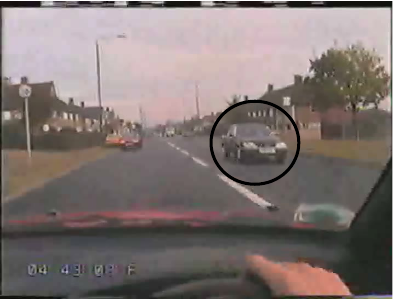}}  \\
  \subfloat[go\_away]{\label{fig:goAway}\includegraphics[width = 40mm]{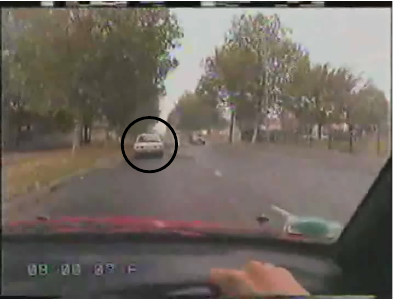}} \qquad  
  \subfloat[overtake]{\label{fig:overtake}\includegraphics[width = 40mm]{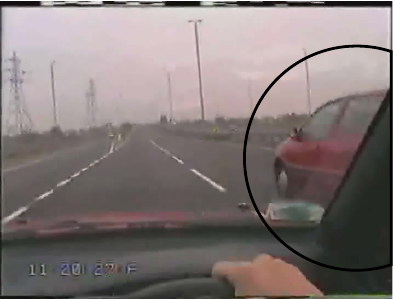}}  \\
  \caption{Attention maps generate by six different methods}  
  \label{fig:SMs}
\end{figure}

\begin{table}[!htbp]
\caption{Normalized Saliency Values in various driving contexts}
\begin{center}
\begin{tabular}{|l|c|c|c|c|c|c|}
\hline
\textbf{Observations} & \multicolumn{1}{l|}{\textbf{ITTI}} & \multicolumn{1}{l|}{\textbf{GBVS}} & \multicolumn{1}{l|}{\textbf{PFT}} & \multicolumn{1}{l|}{\textbf{PQFT}} & \multicolumn{1}{l|}{\textbf{ENT-CON}} & \multicolumn{1}{l|}{\textbf{ENT-KLD}} \\ \hline
\textbf{go\_alone} & 0.01851 & 0.33936 & 0.14633 & 0.13281 & 0.37296 & 0.49269 \\ \hline
\textbf{head\_toward} & 0.02014 & 0.38734 & 0.15944 & 0.16283 & 0.45549 & 0.60256 \\ \hline
\textbf{go\_away} & 0.06923 & 0.36038 & 0.11404 & 0.14979 & 0.28620 & 0.45065 \\ \hline
\textbf{overtake} & 0.07783 & 0.43606 & 0.14036 & 0.19497 & 0.43177 & 0.59839 \\ \hline
\textbf{slow\_down} & 0.02451 & 0.41908 & 0.16767 & 0.16924 & 0.45894 & 0.59269 \\ \hline
\textbf{follow\_car} & 0.03816 & 0.37397 & 0.16606 & 0.16504 & 0.36305 & 0.52260 \\ \hline
\textbf{Mean} & 0.04140 & 0.38603 & 0.14898 & 0.16245 & 0.39473 & 0.54326 \\ \hline
\textbf{Variance} & 0.00056 & 0.00110 & 0.00034 & 0.00036 & 0.00374 & 0.00343 \\ \hline
\end{tabular}
\end{center}
\label{tab:NSVdrc}
\end{table}

\begin{table}[!htbp]
\caption{Difference in Normalized Saliency Values between go\_alone and others }
\begin{center}
\begin{tabular}{|l|c|c|c|c|c|c|}
\hline
\textbf{Observations} & \multicolumn{1}{l|}{\textbf{ITTI}} & \multicolumn{1}{l|}{\textbf{GBVS}} & \multicolumn{1}{l|}{\textbf{PFT}} & \multicolumn{1}{l|}{\textbf{PQFT}} & \multicolumn{1}{l|}{\textbf{ENT-CON}} & \multicolumn{1}{l|}{\textbf{ENT-KLD}} \\ \hline
\textbf{go\_alone} & 0.00000 & 0.00000 & 0.00000 & 0.00000 & 0.00000 & 0.00000 \\ \hline
\textbf{head\_toward} & 0.00146 & 0.04781 & 0.01294 & 0.02982 & 0.01619 & 0.02730 \\ \hline
\textbf{go\_away} & 0.04952 & 0.02009 & -0.03281 & 0.01635 & -0.01624 & -0.00526 \\ \hline
\textbf{overtake} & 0.04721 & 0.04149 & 0.01184 & 0.03407 & 0.00109 & 0.05625 \\ \hline
\textbf{slow\_down} & 0.00996 & 0.02267 & 0.01569 & 0.04101 & 0.01768 & 0.05115 \\ \hline
\textbf{follow\_car} & 0.01574 & 0.04859 & 0.00204 & 0.02750 & -0.00926 & -0.01375 \\ \hline
\textbf{Mean} & 0.02065 & 0.03011 & 0.00162 & 0.02479 & 0.00158 & 0.01928 \\ \hline
\textbf{Variance} & 0.00041 & 0.00031 & 0.00027 & 0.00018 & 0.00015 & 0.00075 \\ \hline
\end{tabular}
\end{center}
\label{tab:dNSVdrc}
\end{table}

Besides the studies in general driving scenes, statistical correlation between saliency maps and human response to specific driving situations are interested as well. It helps answer the question how specific driving situations may affect relations between saliency maps and human eye fixations. Does that correlation get better or worse generally ? Before answering this question, we need to specify available specific driving contexts available in our recorded data, and the number of frames in the situations should be large enough to ensure the relative generality of our experiments. After skimming throughout the database, there are six significant specific driving situations: \texttt{go\_alone, go\_toward, go\_away, slow\_down, overtake and follow\_car}. These situations are named after main tasks which drivers carry out; for example, \texttt{go\_alone, slow\_down and follow\_car} are used to describe situations wherein a car is driven alone without any other moving objects on roads - the figure \ref{fig:goAlone}, is slowed down when reaching a street corner - the figure \ref{fig:slowDown}, and the driver follows a car in front - the figure \ref{fig:followCar}. Likewise, \texttt{go\_toward} or \texttt{go\_away} are used for in-front cars which are moving toward the driver 's car - the figure \ref{fig:goToward}, or moving away the driver's car - the figure \ref{fig:goAway}. Similarly, \texttt{overtake} is named after the situation wherein behind cars suddenly overtake the driver's car - the figure \ref{fig:overtake}. In the \texttt{go\_alone} context, there are no dynamically moving object on roads, so it  should have less visually salient objects and values at eye-marks of those scenes. In other words, the normalized saliency values of \textit{go\_alone} should be less than any other cases. The table \ref{tab:dNSVdrc} summarized differences in normalized saliency values at human fixation points, the table \ref{tab:NSVdrc}, between \texttt{go\_alone} and other driving situations. Almost all values in the table \ref{tab:dNSVdrc} is positive, which means increment in performance of saliency methods whenever drivers are found in particular traffic situations rather than being alone on the roads. Expectingly, the ENT-KLD and ENT-CON saliency method still possesses the best normalized saliency values in the table \ref{tab:NSVdrc}, and its performance in specific driving contexts are generally better than \texttt{go\_alone} situations; there are some decrements in some specific situations because of shortage of frames in that context. More video samples need to be collected and analysed later to clearly find out how saliency maps and human eye-fixations are related in a particular situation. In the constrained number of input video samples and experiments, saliency methods and human psychological data generally have more match in specific driving situations than being alone on the roads.

\begin{figure}[!htbp]
\begin{center}
         \includegraphics[width = \linewidth]{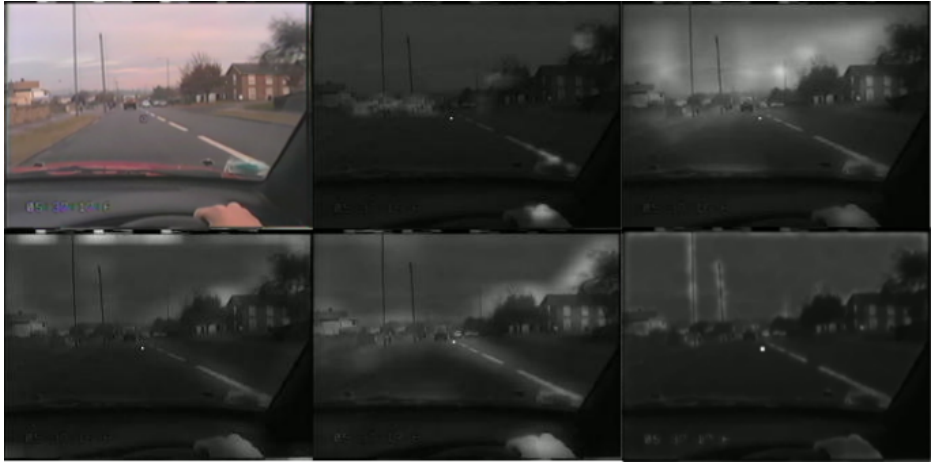}
\end{center}
\caption{Samples and Saliency Maps of ITT, GBVS, PFT, and PQFT, and ENT methods
(left-to-right top-to-bottom order). Please see the supplementary materials for
sample sections of
the video containing the original scene with the eye fixation marker overlayed
on
it and saliency maps of various methods.}
\label{fig:slm3}
\end{figure}

\subsubsection{MSRD Database}
\label{subsubsec:msdrdb}

\begin{figure}[!htbp]
\captionsetup[subfloat]{format=hang,labelfont={rm,bf},textfont={footnotesize},labelformat=simple,labelsep=period,margin=5pt,justification=raggedright}
\centering
\subfloat[Sample Image]{\label{fig:msdr_sample}\includegraphics[width=35mm]{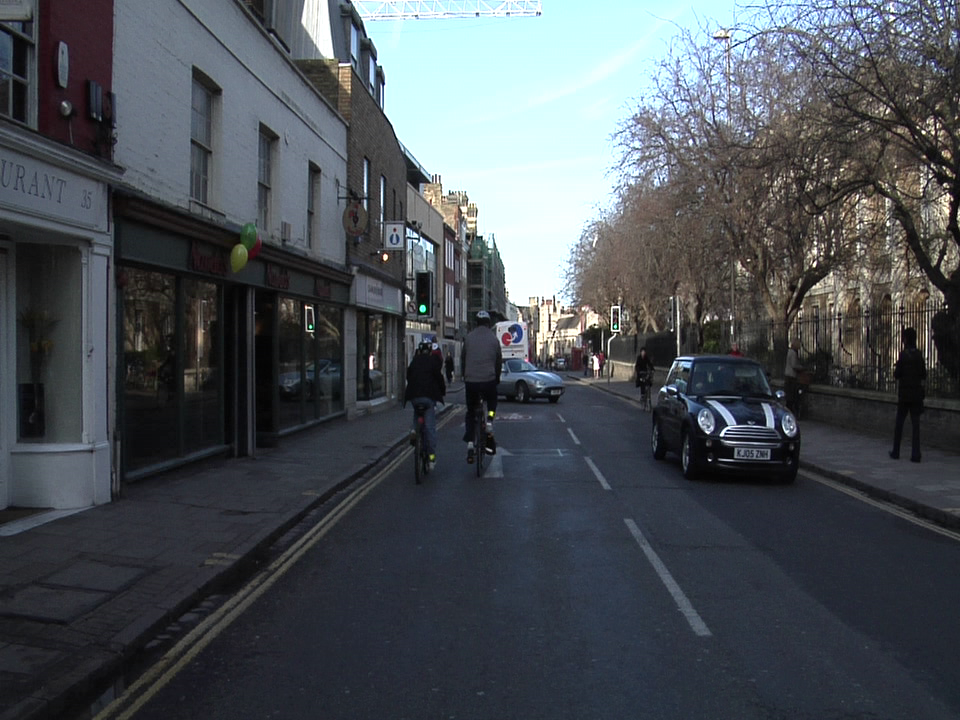}}\qquad
\subfloat[Ground Truth]{\label{fig:msdr_groundtruth}\includegraphics[width=35mm]{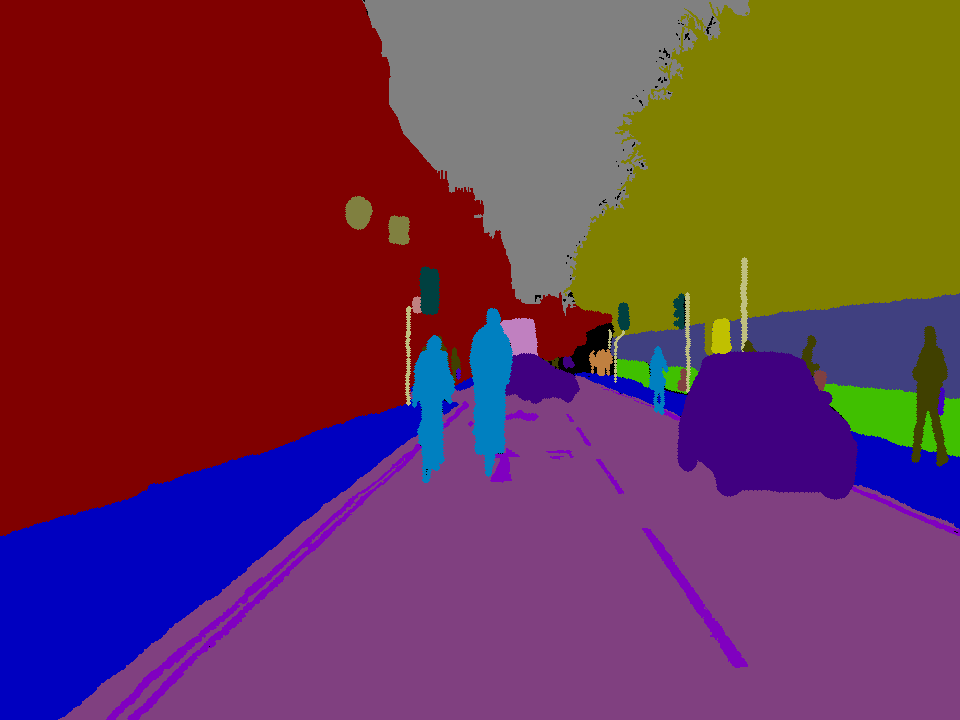}} \qquad
\subfloat[Importance Map]{\label{fig:importance_map}\includegraphics[width=35mm]{msdr_groundtruth.png}}
\caption{MSDR Data Samle Image, Ground Truth}
\label{fig:msdr_database}
\end{figure}

In the previous AUTOM database section, there are a few studies which mainly focus on psychological fitness aspect of computational saliency maps by usage of eye-fixation data. In this section, a study focusing on application aspect of saliency approaches is done on Motion-based Segmentation and Recognition dataset (MSRD) \cite{Brostow2009} with their ground-truth data; figure \ref{fig:msdr_database} reveals a sample of a video from MSDR database with its ground truth.

Before diving into specific details of the study and its evaluation, I need to briefly describe the MSRD data and reasons why it is chosen to be studied. Another name of MSRD database is Cambridge-driving Labeled Video Database (CamVid) which is a collection of driving context videos with object class semantic metadata. It provides groud truth labels that associate each pixel with one of 32 semantic classes. Contrary to majority of traffic database which is recorded from CCTV camera, it uses similar approach to aforementioned AUTOM database which capture video data from the perspective of an automobile driver. Moreover, the driving scenarios are increasingly complex and filled with a large number and heterogeneity of objects in daily driving scenes. Wide range of objects appears in frames is suitable for our study of saliency maps in general driving scenarios. MSRD database provides over ten minutes of high quality 30 Hz footage wherein some parts are semantically labelled at 1Hz and 15Hz. Totally, ground truth data of over 700 images are manually specified by the first human observer and reconfirmed by the second subject for accuracy. Though videos of the whole database last for ten minutes, only 700 sample frames with their correspondent ground truth data are useful in our studies. In addition to 700 frames extracted from original video at 1Hz or 15Hz, four of their past frames are acquired to serve as input data for spatio-temporal saliency maps. 

In MSRD dataset, ground truth data of each sample is a map of manually classified objects wherein each pixel belongs to one of 32 available classes. For a specific application, a certain class of objects will be more important than others. For instance, pedestrians classes bears the most importance values in pedestrian detection applications. Our studies about general driving context and safety leaves us with a certain scale of importance for 32 available classes. Among available classes, there are roughly four groups of objects: the moving objects group, .i.e pedestrian, children, the road objects group, .i.e road, roadshoulder, ceiling objects group, sky, tunnel, and fixed objects group, building, wall, tree. Generally in driving situations, those groups can be arranged in descending order of their importance as moving objects, road objects, fixed objects, and ceiling objects group. Inside each group, object classes are ranked for its importance in the scene; then, we will have importance score for every available object class in the table \ref{objImp}.

\begin{table}[!htbp]
\caption{Importance order of MSRD object classes}
\begin{center}
\begin{tabular}{|l|r|l|l|l|r|}
\hline
\textbf{Moving objects} & \multicolumn{1}{l|}{\textbf{Imp}} & \textbf{Road objects} & \textbf{Imp} & \textbf{Fixed objects} & \multicolumn{1}{l|}{\textbf{Imp}} \\ \hline
Child & 32 & Road & \multicolumn{1}{r|}{18} & TrafficLight & 17 \\ \hline
Pedestrian & 31 & RoadShoulder & \multicolumn{1}{r|}{19} & SignSymbol & 16 \\ \hline
Animal & 30 & LaneMkgsDriv & \multicolumn{1}{r|}{20} & TrafficCone & 15 \\ \hline
Bicyclist & 29 & LaneMkgsNonDriv & \multicolumn{1}{r|}{21} & Column\_Pole & 14 \\ \hline
MotorcycleScooter & 28 & \textbf{Ceiling objects} & \textbf{Imp} & Sidewalk & 13 \\ \hline
CartLuggagePram & 27 & Archway & \multicolumn{1}{r|}{3} & Bridge & 12 \\ \hline
Car & 26 & Tunnel & \multicolumn{1}{r|}{2} & ParkingBlock & 11 \\ \hline
SUVPickupTruck & 25 & Sky & \multicolumn{1}{r|}{1} & Misc\_Text & 10 \\ \hline
Truck\_Bus & 24 &  &  & Building & 9 \\ \hline
Train & 23 &  &  & Fence & 8 \\ \hline
Misc & 22 &  &  & Wall & 7 \\ \hline
 & \multicolumn{1}{l|}{} &  &  & Tree & 6 \\ \hline
 & \multicolumn{1}{l|}{} &  &  & VegetationMisc & 5 \\ \hline
 & \multicolumn{1}{l|}{} &  &  & Void & 4 \\ \hline
\end{tabular}
\end{center}
\label{objImp}
\end{table}

Using the table \ref{objImp}, image pixels of specific classes in the ground truth data can be mapped to its corresponding importance values. In other words, importance maps can be constructed from ground truth data of MSRD dataset. Currently, each pixel in importance maps have integer values in the range of $[ 1,32 ] $, while saliency values are decimal numbers in the range $[0,1]$. Then values of importance maps need scaling down to $[0,1]$ to ensure reliability of further evaluations.

Normalized importance maps shows us how relatively vital each pixel on that map is our purposes, and we also want to know whether higher saliency values in saliency maps means that pixels is more meaningful and necessary to our purposes or not. That leads use to the question how well correlated normalized importance maps and normalized saliency maps are. The normalized cross correlation is chosen to measure and demonstrate that relation. Lets assumed that there are a saliency map $S$ in short of $S(x,y)$ and importance map $M$ in short of $M(x,y)$, the normalized cross correlation $NORMXCORR$ is formulated as the equation \ref{eqn:normxcorr}, and its range of value is $[-1,1]$.

\begin{equation}
NORMXCORR = \frac{\sum_{x,y}[S - \mu(S)][M-\mu(M)]}{(\sum_{x,y}[S-\mu(S)]^2 \sum_{x,y}[M-\mu(M)]^2)^\frac{1}{2}}
\label{eqn:normxcorr}
\end{equation}

\begin{figure}[!htbp]
\begin{center}
         \includegraphics[width = \linewidth]{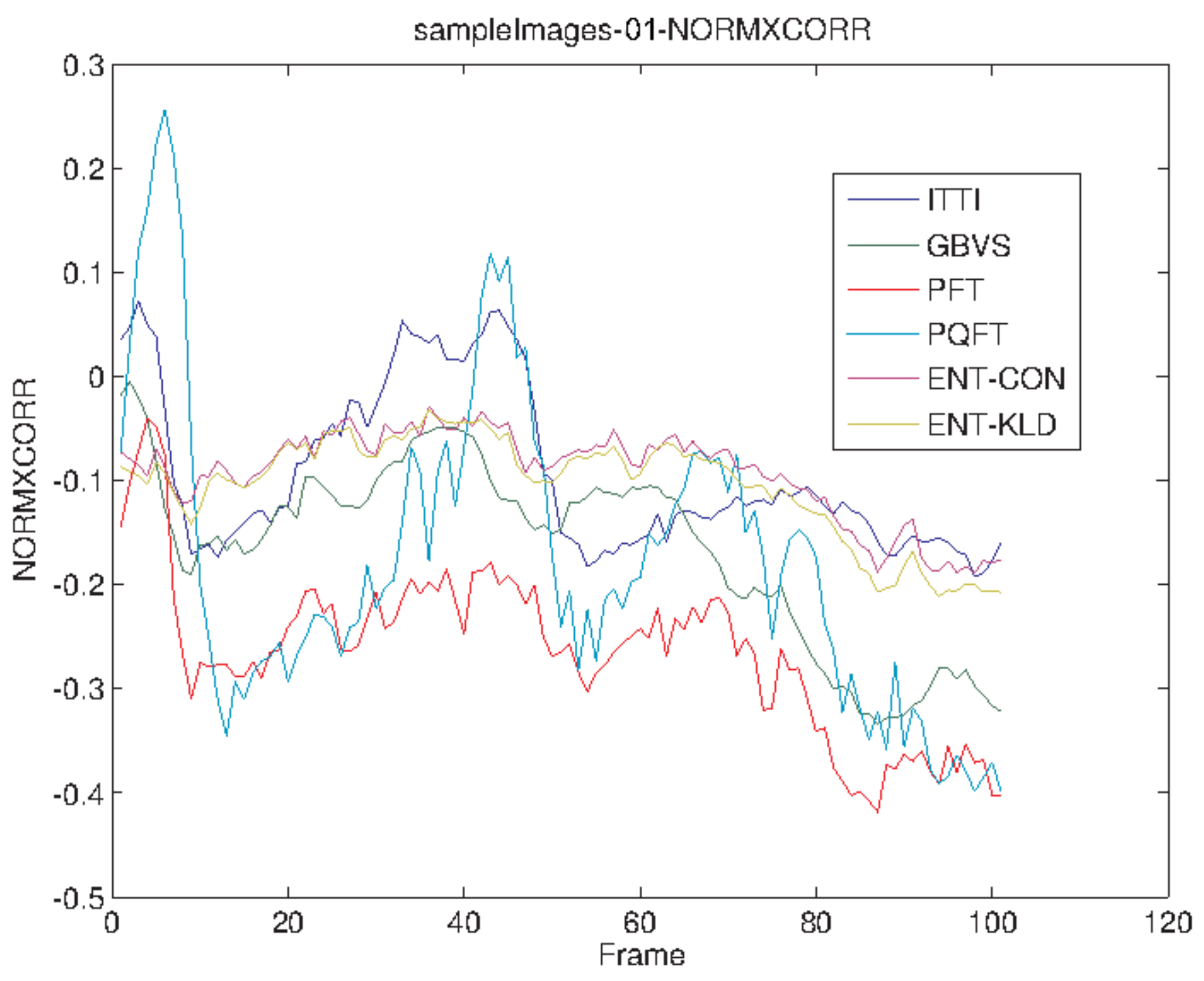}
\end{center}
\caption{Normalized Cross Correlation for sequence\_01}
\label{fig:normxcross1}
\end{figure}

\begin{figure}[!htbp]
\begin{center}
         \includegraphics[width = \linewidth]{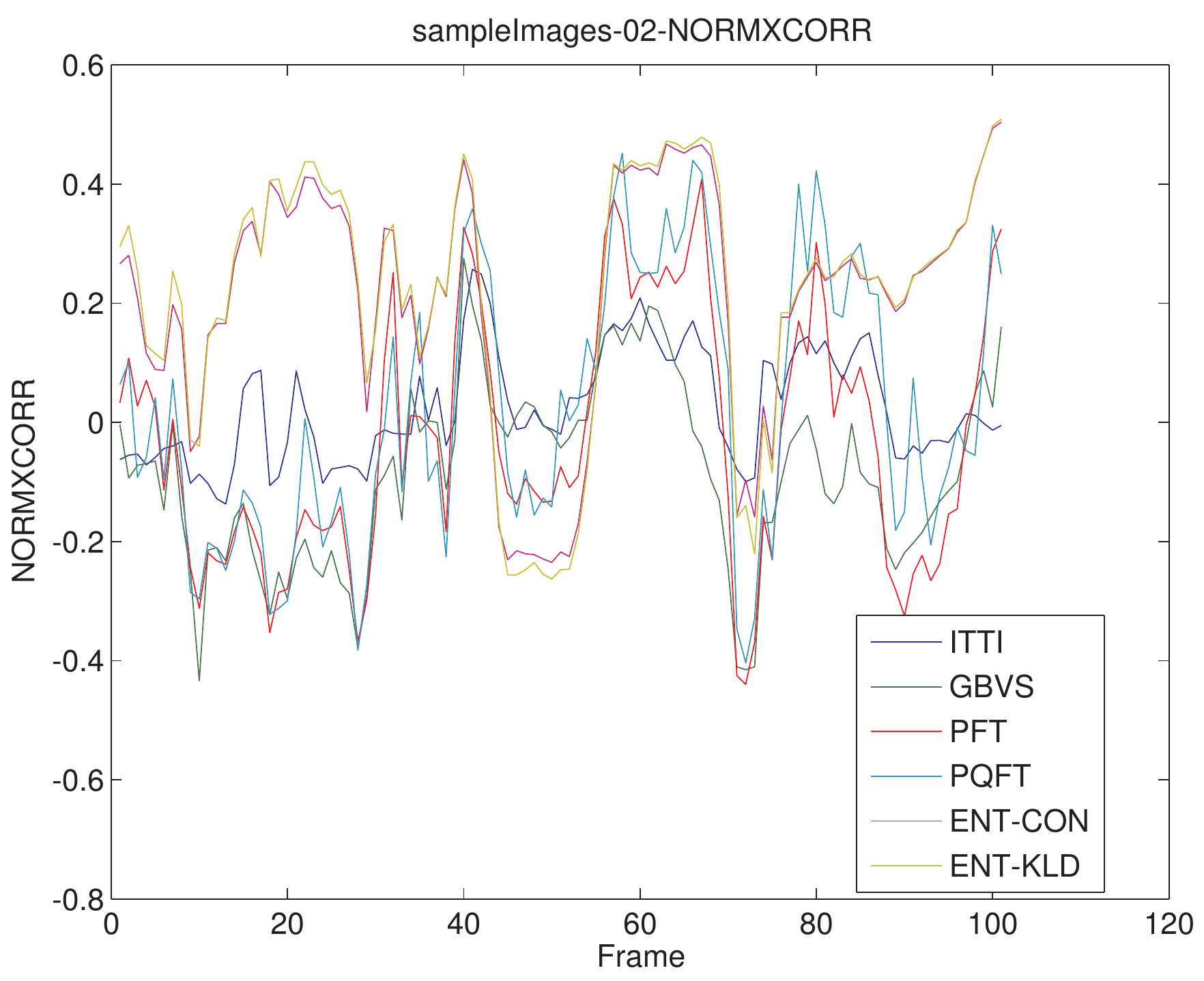}
\end{center}
\caption{Normalized Cross Correlation for sequence\_02}
\label{fig:normxcross2}
\end{figure}

\begin{figure}[!htbp]
\begin{center}
         \includegraphics[width = \linewidth]{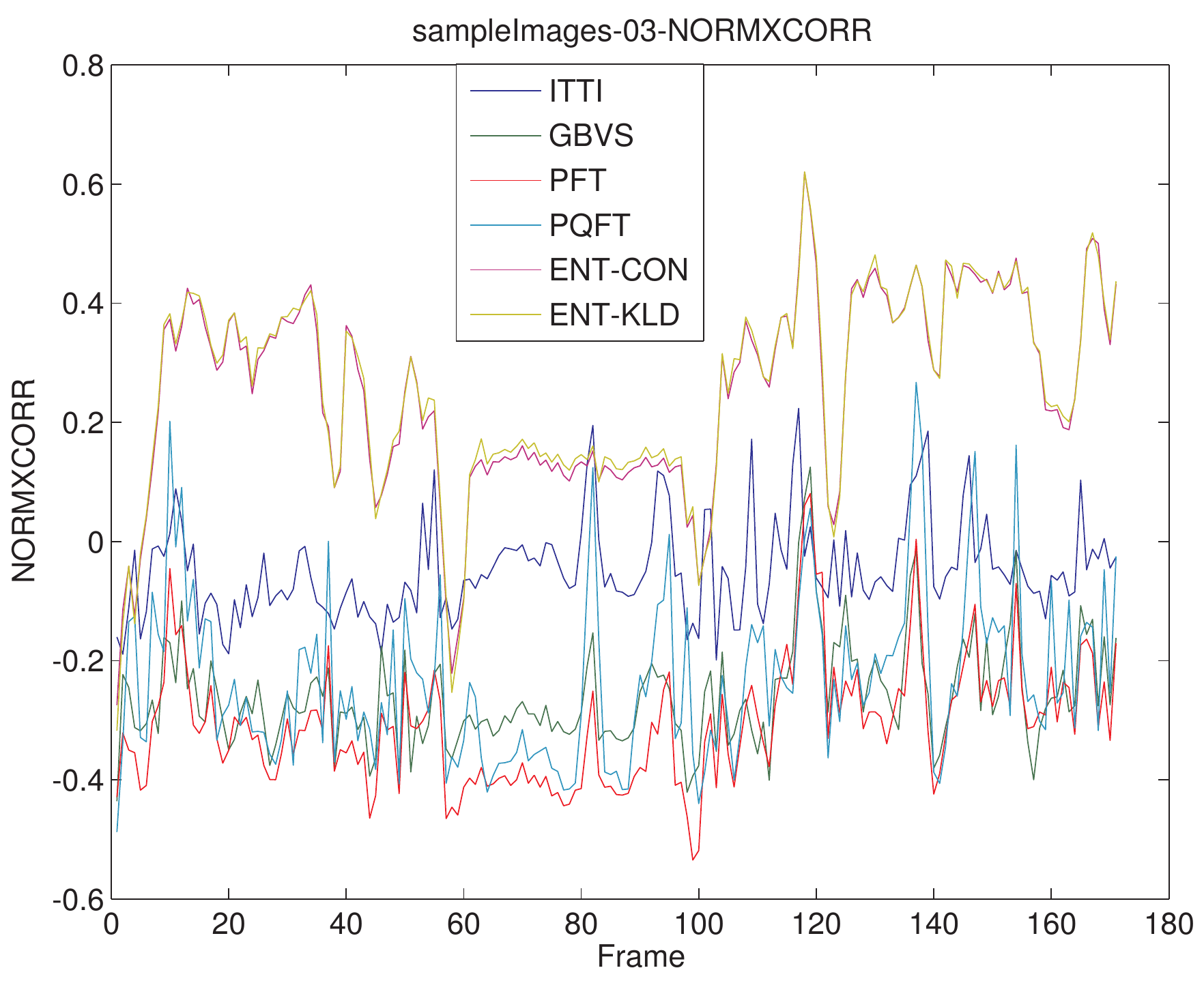}
\end{center}
\caption{Normalized Cross Correlation for sequence\_03}
\label{fig:normxcross3}
\end{figure}

Among over 700 useful data images, they belong to three video scenes. In order to save space and generalize the simulation results, NORMXCORR of frames from three video sequences will be plotted in three figures: figure \ref{fig:normxcross1},\ref{fig:normxcross2},\ref{fig:normxcross3}. Each figure has six lines with different colour representing NORMXCROSS between importance maps and saliency maps from six saliency methods over frames of a sequence. The plots show how the measurement fluctuate and change across the whole video, and their statistical mean and variance are also displayed in the tables \ref{tab:normxcross1}, \ref{tab:normxcross2}, \ref{tab:normxcross3}.

\begin{table}[!htbp]
\caption{Statistical data of Normalized Cross Correlation over video sequence 01}
\begin{center}
\begin{tabular}{|c|c|c|c|c|c|c|}
\hline
\textbf{Observations} & \textbf{ITTI} & \textbf{GBVS} & \textbf{PFT} & \textbf{PQFT} & \textbf{INFO} & \textbf{ENTRO} \\ \hline
\textbf{Mean} & -0.09315 & -0.16580 & -0.26443 & -0.17535 & -0.09301 & -0.10431 \\ \hline
\textbf{Variance} & 0.00615 & 0.00775 & 0.00620 & 0.02255 & 0.00188 & 0.00248 \\ \hline
\end{tabular}
\end{center}
\label{tab:normxcross1}
\end{table}

\begin{table}[!htbp]
\caption{Statistical data of Normalized Cross Correlation over video sequence 02}
\begin{center}
\begin{tabular}{|c|c|c|c|c|c|c|}
\hline
\textbf{Observations} & \textbf{ITTI} & \textbf{GBVS} & \textbf{PFT} & \textbf{PQFT} & \textbf{INFO} & \textbf{ENTRO} \\ \hline
\textbf{Mean} & 0.02512 & -0.08502 & -0.03021 & 0.02062 & 0.20900 & 0.21499 \\ \hline
\textbf{Variance} & 0.00866 & 0.02305 & 0.04298 & 0.04906 & 0.04088 & 0.04544 \\ \hline
\end{tabular}
\end{center}
\label{tab:normxcross2}
\end{table}

\begin{table}[!htbp]
\caption{Statistical data of Normalized Cross Correlation over video sequence 03}
\begin{center}
\begin{tabular}{|c|c|c|c|c|c|c|}
\hline
\textbf{Observations} & \textbf{ITTI} & \textbf{GBVS} & \textbf{PFT} & \textbf{PQFT} & \textbf{INFO} & \textbf{ENTRO} \\ \hline
\textbf{Mean} & -0.04801 & -0.26006 & -0.30763 & -0.23102 & 0.24872 & 0.25500 \\ \hline
\textbf{Variance} & 0.00638 & 0.00763 & 0.01083 & 0.01970 & 0.02723 & 0.02761 \\ \hline
\end{tabular}
\end{center}
\label{tab:normxcross3}
\end{table}

After looking through three plots and three tables, NORMXCROSS of the proposed method is larger than the other established methods. Though in some psychological measurements, the proposed ENT method does not give the best performance, its saliency maps are well correlated with application oriented importance maps; in other words, it provides clues about how important and meaningful each pixel is to the scene.

\section{Concluding Remarks}
\label{sec:conclusion}

In this paper, we have formulated bottom-up visual saliency as center surround
conditional entropy and presented a fast and efficient computational method for
computing saliency maps on still images and full motion videos. We have
shown that the new method is not only computationally fast and efficient but
also gives state of the art performances on publicly available eye tracking
databases. Furthermore, the technique is feasibly extended from still image saliency, spatial saliency, to video data, spatial temporal saliency, and its efficiency is evaluated on two driving context video database AUTOM and MSRD. As visual saliency is a bottom-up as well as a top-down process, our
future work will investigate the including of top-down context in the
estimation of visual saliency and explore its various applications in various
cognitive researches.

\bibliographystyle{spmpsci} 
\bibliography{JCC2011}

\end{document}